\PassOptionsToPackage{table,dvipsnames}{xcolor}
\documentclass{article} 
\usepackage[final]{colm2026_conference}

\usepackage{microtype}
\usepackage{hyperref}
\usepackage{url}
\usepackage{booktabs}
\usepackage{graphicx}
\usepackage[inkscapearea=page]{svg}
\usepackage{adjustbox}
\usepackage{enumitem}
\usepackage[frozencache,cachedir=.]{minted}
\usepackage{tcolorbox}
\usepackage{gensymb}
\usepackage{amsmath}
\usepackage{minted}
\usepackage{algorithm}
\usepackage{algpseudocode}
\usepackage{subcaption}
\usepackage{xspace}
\usepackage{wrapfig}
\usepackage{makecell}
\usepackage{siunitx}

\setcellgapes{4pt}

\usepackage{lineno}

\definecolor{darkblue}{rgb}{0, 0, 0.5}
\hypersetup{colorlinks=true, citecolor=darkblue, linkcolor=darkblue, urlcolor=darkblue}

\newcommand{\InfiniteScienceGym}{{\textsf{InfiniteScienceGym}}\xspace}

\title{
  \raisebox{-0.6\baselineskip}{\includegraphics[width=0.15\linewidth]{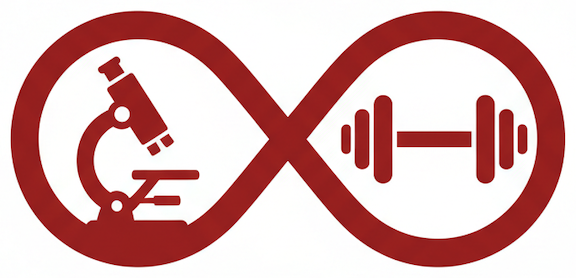}}
  \hspace{0.5em}
  \begin{minipage}{0.8\linewidth}
    \InfiniteScienceGym: An Unbounded, Procedurally-Generated Benchmark for Scientific Analysis
  \end{minipage}
}

\definecolor{ISGred}{RGB}{137, 42, 36}
\author{Oliver Bentham \& Vivek Srikumar\\
    Kahlert School of Computing \\
    University of Utah \\
    \texttt{oliver.bentham@utah.edu}\quad\texttt{svivek@cs.utah.edu} \\
    \raisebox{-1.5pt}{\includegraphics[height=1em]{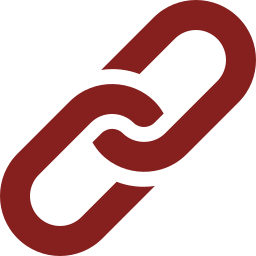}}\;\href{https://infinitesciencegym.github.io/}{\textcolor{ISGred}{\textbf{https://infinitesciencegym.github.io/}}}
}

\newif\ifshowannotations
\showannotationstrue 
\ifshowannotations
  \newcommand{\authorNote}[4]{\noindent\textbf{\textcolor{#1}{[$^{#2}_{#3}$: \textit{#4}]}}}
\else
  \newcommand{\authorNote}[4]{}
\fi

\begin{document}

\ifcolmsubmission
\linenumbers
\fi

\maketitle

\begin{abstract}
Large language models are emerging as scientific assistants, but evaluating their ability to reason from empirical data remains challenging.
Benchmarks derived from published studies and human annotations inherit publication bias, known-knowledge bias, label noise, and substantial storage requirements.
We present \InfiniteScienceGym, a procedurally generated benchmark of scientific repositories paired with a verifiable question-answering task.
From a seed, the simulator deterministically generates a self-contained repository with realistic directory structure, files, and tabular data, and a privileged QA generator produces both answerable and unanswerable questions with exact ground truth.
This makes it possible to evaluate evidence-grounded reasoning, abstention, and tool-mediated analysis in a controlled setting without distributing a large static corpus.
\InfiniteScienceGym complements real scientific benchmarks by targeting blind spots and failure modes that are hard to evaluate using published datasets alone.
Evaluating both proprietary and open-weight models, we find that none achieve more than 50\% accuracy overall, that recognizing unanswerable questions remains a major weakness, and that stronger models tend to use tools more effectively rather than simply consuming more tokens.
\end{abstract}

\section{Introduction}
\label{sec:introduction}

Large language models (LLMs) are increasingly deployed as scientific assistants, with recent systems contributing to automated research workflows, mathematical reasoning and discovery-oriented tasks \citep{novikov2025alphaevolvecodingagentscientific, feng2026autonomousmathematicsresearch, guevara2026singleminusgluontreeamplitudes}.
Such \emph{AI-for-science} agents are expected to span the gamut of the scientific process and have shown promise for hypothesis generation \citep{agarwal2025autodiscovery}, literature review \citep{tang-etal-2025-large}, and data exploration \citep{chen2025scienceagentbench}.
As both autonomy and scientific stakes get higher, two questions become pressing: \emph{Do our AI-for-science agents reason about available data? Can they recognize when that data is insufficient to support a conclusion?}

\begin{figure}[t]
    \centering
    \vspace{-1em}
    \includegraphics[width=\textwidth, page=1]{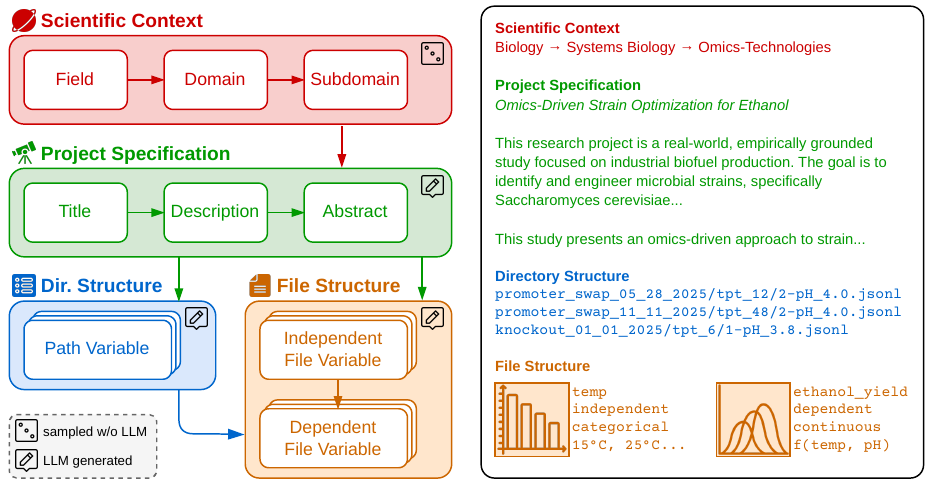}
    \caption{
        Data repositories are generated in a top-down manner.
        1) \textbf{Scientific Context}: A field, domain, and subdomain are randomly sampled from a hierarchical taxonomy.
        2) \textbf{Project Specification}: A project title, description (containing a plan, hypotheses, variables, and confounders), and abstract are generated with the LLM, conditioned on the scientific context.
        3) \textbf{Directory Structure}: The project specification is used to create a plausible directory structure in which folder and file names describe the variables.
        4) \textbf{File Structure}: Conditioned on all prior information, independent and dependent variables are generated using named distributions and functions of other variables, respectively.
        The right side shows one such instantiation, corresponding to seed $118$.
    }
    \label{fig:diagram}
    \vspace{-0.5em}
\end{figure}

Answering these questions requires benchmarks that test empirical data reasoning, but existing evaluations introduce methodological confounders.
First, published research is not a random sample of scientific inquiry; it disproportionately contains positive and ``clean'' findings~\citep{Dickersin1993-ij, Nissen2016-fe}, underrepresenting settings where the correct conclusion is that a question is unanswerable (e.g., missing measurements or mismatched operational definitions).
Second, the reliance on established work entangles data-driven inference with a model's parametric priors.
Models may discount novel or counterintuitive conclusions that conflict with familiar narratives. 
Such known-knowledge bias is hard to isolate with benchmarks derived from existing knowledge.
Third, human annotation in real-data benchmarks introduces label noise and ambiguous ground truth.
Finally, storing and distributing large scientific corpora can be costly and legally constrained.

These limitations suggest a gap.
We need evaluation tools that can control which scientific phenomena are present (or absent), provide verifiable ground truth for both answerable and unanswerable questions, scale without requiring large static datasets, and support agentic behaviors such as tool use and evidence-backed reporting.

To address this need, we introduce \InfiniteScienceGym, a procedurally generated benchmark for scientific data exploration and question answering.\footnote{Code is available at: \href{https://github.com/utahnlp/infinite-science-gym}{https://github.com/utahnlp/infinite-science-gym}}
It consists of three core components.
\begin{enumerate}[nosep]
    \item A \textbf{simulator} that, given a random seed, generates a self-contained scientific repository: a file-system whose contents resemble the artifacts of a real research project (e.g., datasets, metadata, and notes).
    Repositories are deterministic functions of the seed: repository \#1 is always reproduced by \verb|seed=1|.
    Since a repository is generated on the fly from its seed, \InfiniteScienceGym avoids the storage burden of distributing large static corpora while providing effectively unbounded evaluation instances.
    \item A \textbf{question-answer (QA) generator} that uses privileged access to the simulator's underlying process to produce templated questions with guaranteed ground truth, including both answerable and unanswerable questions.
    \item A \textbf{paraphrase module} that converts templated QA pairs into more naturalistic research queries, allowing us to probe robustness to surface-form variation while retaining verifiability.
\end{enumerate}

Together, these components provide a benchmark with exact ground truth, provable unanswerability, naturalistic surface form and unlimited scale without structural contamination with existing models or benchmarks.

We study three questions about data-centric scientific assistants using \InfiniteScienceGym:
\begin{enumerate}[label=\textbf{RQ\arabic*:}, nosep]
  \item How do different LLMs compare in their ability to correctly answer questions grounded in the repository data?
  \item Do models respond appropriately when presented with unanswerable questions?
  \item How effectively do models interact with data and tools to support their answers?
\end{enumerate}

We evaluate proprietary and open-weight models using a tool-enabled setup that mirrors the intended use of scientific assistants: models may inspect repository contents and invoke analysis when needed. 
We find that none of the evaluated models perform well overall on \InfiniteScienceGym, with the best model achieving only 49.2\% accuracy.
We also find that identifying unanswerable questions remains a weakness: many systems provide confident answers even when the repository does not contain sufficient evidence.
Looking into different problem-solving strategies, we discover that performance scales with interaction, that accuracy increases with the number of tool calls made during an episode.
By contrast, token usage is a misleading proxy for ``effort'' or ``thoroughness''.
Stronger agents often avoid loading large files into the context window and instead operate on them programmatically, efficiently analyzing the data without a proportional increase in tokens.

In summary, this paper makes the following contributions:
\begin{itemize}[nosep]
  \item We introduce \InfiniteScienceGym, a procedurally generated benchmark of scientific repositories paired with verifiable QA tasks, including unanswerable questions.
  \item We describe a practical generation pipeline (simulator $\rightarrow$ QA generator $\rightarrow$ paraphrase module) that produces reproducible, seed-indexed repositories and ground-truth QA pairs without storing large static datasets.
  \item We provide an evaluation of both proprietary and open-weight models, demonstrating systematic failures on unanswerable questions and showing that accuracy increases with tool interaction, and that token usage alone can mischaracterize how much evidence a model has processed.
  \item We position \InfiniteScienceGym as a complement to real-data benchmarks: it is designed to expose controlled blind spots and failure modes that are hard to evaluate using published datasets alone.
\end{itemize}

\section{The Case for Procedurally Generated Scientific Evaluation}
\label{sec:background}

\InfiniteScienceGym sits at the intersection of research on benchmarks for data-driven scientific reasoning, evaluation of abstention and uncertainty under controlled or simulated conditions, and agentic reasoning over structured artifacts.

Several recent benchmarks evaluate LLMs and agents on scientific reasoning grounded in empirical data.
DiscoveryBench~\citep{majumder2025discoverybench} formalizes multi-step discovery workflows, ScienceAgentBench~\citep{chen2025scienceagentbench} evaluates language agents on tasks extracted from peer-reviewed scientific work, AutoDiscovery~\citep{agarwal2025autodiscovery} studies open-ended discovery via Bayesian surprise, and LLM-SRBench~\citep{shojaee2025llmsrbench} focuses on scientific equation discovery.
These are largely grounded in curated real datasets, published studies, or discovery settings where success is defined by finding a valid result.
Consequently, they are less suited for isolating cases where the correct behavior is to conclude that the available evidence is insufficient.
\InfiniteScienceGym is designed to complement this literature by procedurally generating repository-grounded scientific tasks in which answerability and unanswerability are known by construction.

A second relevant line of work studies abstention, uncertainty, and evaluation under controlled novelty.
Prior work argues that LLMs are often rewarded for guessing rather than acknowledging uncertainty~\citep{kalai2025languagemodelshallucinate}. A growing literature studies how to detect knowledge gaps and induce abstention~\citep[e.g.,][]{feng-etal-2024-dont, wen-etal-2025-know, xia-etal-2025-survey}.
ALCUNA~\citep{yin2023alcuna} uses synthetic entities and altered relations to test models under new knowledge that may conflict with prior beliefs, while SimulBench~\citep{jia-etal-2025-simulbench} shows how simulation-based benchmarks can support controlled, interactive evaluation.
\InfiniteScienceGym is aligned with this perspective with a controlled setting where abstention, underdetermination, and evidence-grounded reasoning can be evaluated directly.

A third related line of work evaluates models on structured artifacts such as code, tables, and databases.
SciCode~\citep{tian2024scicode} and CORE-Bench~\citep{corebench} study scientific workflows through research coding and computational reproducibility tasks grounded in real artifacts.
Related coding benchmarks include SWE-bench~\citep{jimenez2024swebench}, which evaluates GitHub repository issue resolution, and RepoBench~\citep{liu2023repobench}, which targets code retrieval and multi-file code completion.
Adjacent work on structured-data reasoning includes BIRD~\citep{li2024can} and Spider 2.0~\citep{lei2025spider} for database-grounded text-to-SQL, and several benchmarks focusing on reasoning over semi-structured tables~\citep[][inter alia]{TableBench,Chen2020TabFact,gupta-etal-2020-infotabs}.
These benchmarks move beyond free-form text and require multi-step reasoning over structured evidence, but none combines file system navigation, tool-mediated analysis, and verifiable unanswerability.

These benchmark families highlight a set of competing desiderata for evaluating scientific assistants: fidelity to scientific workflows, verifiable ground truth, scalable evaluation, robustness to publication and known-knowledge biases, and support for agentic setups for file navigation and tool use.
Existing benchmarks satisfy different subsets of these goals.
\InfiniteScienceGym occupies this intersection and prioritizes seed-based reproducibility and exact ground-truth verifiability at effectively unbounded scale.

\section{\InfiniteScienceGym: Reaping Repositories from Sown Seeds}
\label{sec:infinite_science_gym}

\InfiniteScienceGym consists of a simulator, a QA generator, and a paraphrase module.
Given a random seed, the simulator produces complete file systems and their contents.
Using privileged access to the data-generating process, the QA generator creates answerable and unanswerable templated questions.
Finally, the paraphrase module uses privileged metadata to reword the questions as naturalistic research queries.

All stochastic choices are deterministically derived from a given random seed $i$, ensuring that repository $\#i$ is always identical across runs.
By design, the number of file systems and QA pairs produced is unbounded, and reproducibility is guaranteed.

\subsection{Simulating Scientific Repositories}

We construct repositories in a top-down manner, beginning with a randomly chosen scientific area and iteratively defining the project context, directory structure and path variables, and, finally, the tabular file structure (Figure~\ref{fig:diagram}).
Except for the scientific context, each step requires an LLM to generate its output.
In our simulator, we use Qwen3 4B Instruct.
Each generation step is conditioned on information determined in prior steps, meaning the project and repository begin general and get progressively more detailed.

\paragraph{Scientific context.}
We first sample a field, domain, and subdomain from a hierarchical taxonomy broadly covering empirical science.
The taxonomy contains 22 fields, 244 domains, and 780 subdomains, described in more detail in Appendix~\ref{sec:scientific_taxonomy}.
An example of a hierarchical triple from the taxonomy is ``Computer Science''$\rightarrow$``Artificial Intelligence and Machine Learning''$\rightarrow$``AI Safety and Robustness''.
Conditioning all downstream generation on a specific subdomain encourages more coherent and diverse project ideas.

\paragraph{Project specification.}
Conditioned on the sampled scientific context, we prompt the LLM for $k$ candidate research project titles, following prior findings that prompting for $k$ results produces more diverse outputs than prompting $k$ times \citep{troshin-etal-2025-asking}.
A project title is then sampled from the generated candidate list.
Conditioned on the scientific context and project title, a verbose description is generated that includes a high-level hypothesis, a set of independent variables, dependent variables, potential confounders, and a natural-language description of the experimental setup.
This specification acts as a latent ``data-generating program'' that governs all subsequent steps but is never exposed to a model being evaluated.
We also generate a concise abstract summarizing the project, mimicking the style of a scientific paper. 
While this approach does not yield particularly innovative ideas, it helps produce diverse contexts that cover many scientific areas.
Appendix~\ref{sec:full_example} shows the full project specification corresponding to repository \#118.

\paragraph{Directory structure.}
From the project specification, we generate a directory tree consisting of folders and files that resemble a real research repository.

Paths are defined by a sequence of placeholder variables, which might encode the researcher, the date, or a variable specific to the project specification.
From this perspective, a full file path is nothing more than a series of populated placeholder variables separated by delimiters ``/'', ``\_'', and ``-''.
We generate the templated directory and path structure by using the LLM to iteratively select placeholder variables and delimiters, conditioned on the project specification, until the path template contains $n_\text{path}$ variables.
Path variables correspond to independent variables, and their values encode experimental conditions.
This design requires models to interpret both file contents and file organization when reasoning about the data.
Finally, a file extension is uniformly sampled from $\{\text{csv}, \text{json}, \text{jsonl}, \text{xlsx}, \text{txt}, \text{log}\}$.
Appendix~\ref{sec:cross_product} gives details about deciding which paths are included in the directory tree.

The example below shows the templated directory structure chosen for repository \#118, as well as an example path from that repository.

\begin{tcolorbox}[height=1.2cm, valign=center, colback=white, colframe=black!30,]
\texttt{./\{var1\}/\{var2\}\_\{date\}/\{var3\}/\{seq\_number\}-\{var4\}.jsonl} \\
\texttt{./gphase\_early/gtype\_knockout\_01\_05\_2025/tpt\_0/2-pH\_4.0.jsonl}
\end{tcolorbox}

\begin{table}[t]
    \vspace{-1em}
    \centering
    \begin{tabular}{l|S[table-format=2.1(1)]|S[table-format=2.1(1)] S[table-format=2.1(1)] S[table-format=2.1(1)]}
    \toprule
    &  & \multicolumn{3}{c}{\textbf{Paraphrase Model}} \\
    \textbf{Evaluated Model} & {Templated} & {GPT-OSS 20B} & {Gemma 3 27B it} & {Qwen3 4B Inst.} \\
    \midrule
    GPT-5.4 & 49.2(2) & 48.0(2) & 39.1(2) & 43.0(2) \\
    Claude Opus 4.6 & 43.7(2) & 41.9(2) & 27.8(2) & 28.7(2) \\
    GPT-OSS 20B & 29.7(2) & 30.0(2) & 26.9(2) & 29.7(2) \\
    Qwen3 4B Inst. & 26.5(2) & 24.2(2) & 22.3(2) & 25.5(2) \\
    Gemma 3 27B it & 24.2(2) & 23.2(2) & 22.7(2) & 22.2(2) \\
    \bottomrule
    \end{tabular}
    \caption{Accuracy by evaluated model and template/paraphrase variant. Each score also shows the corresponding 95\% confidence interval. We do not report results averaged across paraphrases; our analysis in Appendix~\ref{sec:paraphrase_validity} shows that paraphrase quality varies significantly across the paraphrasing models, and averaging can be misleading.}
    \label{tab:disaggregated_results}
    \vspace{-0.5em}
\end{table}

\paragraph{File structure.}
Each file contains tabular data, with columns serving one of four roles: identifiers, dates/times, independent variables, or dependent variables.
Given the project specification and path variables, we prompt the LLM to generate a list of relevant file variables, along with their role and, where applicable, their type (categorical, discrete integer, or continuous).
Once the file variables are decided, the LLM generates parameters for each independent variable, depending on its type:
\begin{itemize}[nosep]
    \item \textbf{categorical}: the model produces the discrete list of values and their probabilities.
    \item \textbf{discrete integer}: the model selects a valid distribution from Bernoulli, Binomial, Geometric, Negative Binomial, or Poisson, along with plausible parameters.
    \item \textbf{continuous}: the model chooses a distribution from Beta, Exponential, Normal, or Uniform, along with plausible parameters.
\end{itemize}

With the identifier, datetime, and independent variable columns defined and parameterized, the LLM is prompted to define plausible Python functions for the dependent variables.
Each function takes all prior variables, including the path variables.
These functions may encode linear or nonlinear relationships, noise processes, and partial observability.
An example function is shown in Appendix~\ref{sec:dependent_variables}.

To populate a given file, specified by a repository seed and a file path, we first generate the data repository (scientific context, project specification, path variables, and the file variables).
We then hash the file path into a new seed, set that seed globally, sample the file's row count from a predetermined distribution, and then sample values for the file variables until we reach the target number of rows.
Finally, the tabular file contents are serialized according to the specified file extension.
This algorithm is explained in more detail in Appendix~\ref{sec:populating_a_file}.

To evaluate the plausibility of \InfiniteScienceGym, we perform a human study with 19 domain experts, covering 6/22 fields in the taxonomy.
We find statistically significant evidence that the generated repositories are plausible across all dimensions of interest: repository organization, naming conventions, schema plausibility, value plausibility, and visible simulation artifacts.
This study is discussed in more detail in Appendix~\ref{sec:human_study}.

\subsection{Templated QA Generation with Privileged Simulator Access}
\label{sec:templated_qa_generation}

Given a simulated repository, the QA generator produces question-answer pairs using privileged access to the underlying data-generating process.

Every question is specified by code that samples filtering conditions and converts them into templated language.
If the question is answerable, the QA generator computes the ground-truth answer by performing the appropriate calculations.
The example below shows one such templated question.
This is a difficult question because it has three filters on the directory and file path, and two filters on the file variables.

\begin{tcolorbox}[height=2cm, valign=center, colback=white, colframe=black!30]
    Only considering files where the ``seq\_number'' is ``4'', the ``date'' is ``2025-05-28'', and the ``pH'' is one of ``4.0'', ``5.0'', or ``6.0'', and only considering rows where the ``temp'' is ``35'' and the ``biomass'' is in the range 0.17-0.465, what is the median value of the ``residual\_glucose'' variable?
\end{tcolorbox}

The QA generator can produce both answerable and unanswerable questions.
Answerable questions correspond to queries with uniquely determined answers from the repository contents (e.g., queries about statistical relationships, comparisons across conditions, or aggregations).
Unanswerable questions are constructed such that the available data is insufficient to determine the answer
due to missing variables, unobserved confounders, insufficient statistical power, or ambiguity in operational definitions.
For example, the question above would be unanswerable if the filters produced no files or rows on which to calculate a median.
Another way the question might be unanswerable is if the ``residual\_glucose'' variable is not continuous, and calculating its median is not a valid operation.

Appendix~\ref{sec:question_templates} lists the types of templated questions, along with examples.

\begin{figure}[t]
    \centering
    \vspace{-1em}
    \includegraphics[width=\linewidth]{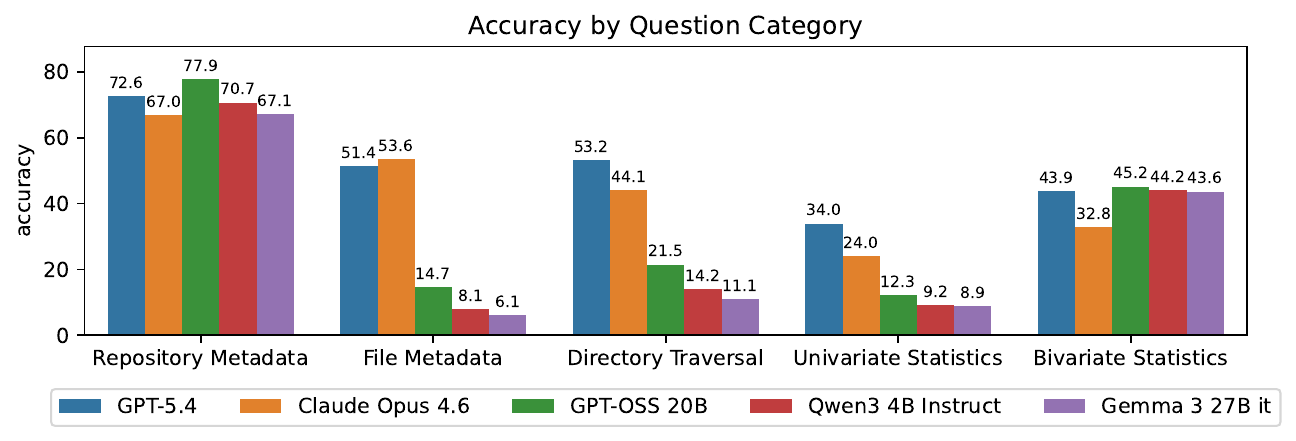}
    \caption{Model accuracy, separated by question category.}
    \label{fig:accuracy_by_question_category}
    \vspace{-0.5em}
\end{figure}

\subsection{From Templated to Naturalistic Questions}

While templated questions provide control and verifiability, they are unnatural in style.
To bridge this gap, the paraphrase module converts templated questions into more naturalistic language.
It conditions on the full project specification, including the project description and all variable descriptions, to produce queries that resemble how a researcher might formulate a question.
The example below shows the same templated example from \S~\ref{sec:templated_qa_generation}, paraphrased using Gemma 3 27B it \citep{gemma_2025}.

\begin{tcolorbox}[height=2.4cm, valign=center, colback=white, colframe=black!30]
    For the fourth experimental run conducted on May 28th, 2025, considering only those conditions where the fermentation medium's acidity ranged from 4.0 to 6.0 and the fermentation temperature was 35 degrees Celsius, what is the median amount of glucose remaining after fermentation when the biomass concentration was between 0.17 and 0.465 grams per liter?
\end{tcolorbox}

More examples of paraphrased questions can be found in Appendix~\ref{sec:paraphrase_examples}.
\section{Experimental Setup}
\label{sec:experiments}

We conduct two experiments using \InfiniteScienceGym.
The first experiment addresses the research questions in \S~\ref{sec:introduction} by considering models' responses to a sample of templated and paraphrased QA pairs covering answerable and unanswerable questions.
We compare models' overall performance, their ability to detect unanswerable questions, and their token and tool usage.
We also manually annotate a subset of incorrect answers from each model to identify error patterns that illustrate model shortcomings.
The second experiment examines robustness to surface-form variation by measuring agreement between model responses to templated and paraphrased variants of each question.
This experiment also includes a manual audit of template/paraphrase question pairs to determine which paraphrases introduce warranted difficulty and which add unhelpful noise.

We sample 500 questions from the 15,988 questions corresponding to the first 500 repositories to encourage diversity across scientific domains and simulated projects. 

All simulated data repositories are generated using Qwen3 4B Instruct \citep{qwen3technicalreport}.
We use the capable open-weight models GPT-OSS 20B \citep{openai2025gptoss120bgptoss20bmodel} and Gemma 3 27B it as paraphrase models, along with Qwen3 4B Instruct, which generated the repositories.
We evaluate OpenAI GPT-5.4, Anthropic Claude Opus 4.6, GPT-OSS 20B, Gemma 3 27B it, and Qwen3 4B Instruct.

See Appendix~\ref{sec:questions_in_evaluation} for details about the questions and Appendix~\ref{sec:eval_details} for evaluation details, including the grading rubrics for different question types.

\section{Experimental Results and Analysis}
\label{sec:results}

\begin{figure}[t]
    \centering
    \vspace{-1em}
    \includegraphics[width=\linewidth]{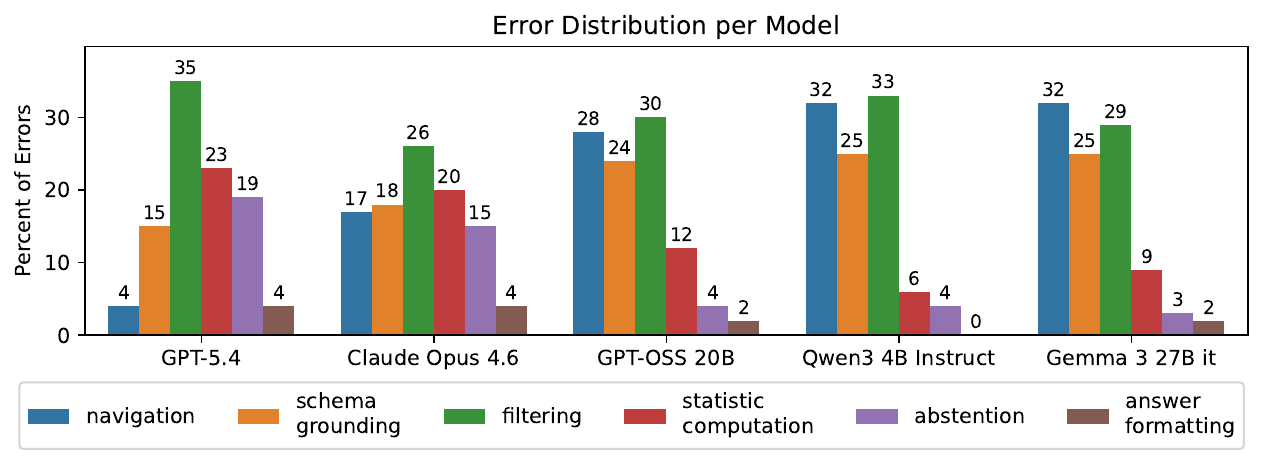}
    \caption{Distribution of error types by model, based on samples of 100 incorrect QA pairs per model (paraphrased using GPT-OSS 20B). Distributions are normalized within each model. Note that the error categories have a sequential structure. See the text for details.}
    \label{fig:errors}
    \vspace{-0.5em}
\end{figure}

\paragraph{How do models perform on \InfiniteScienceGym?}
Table~\ref{tab:disaggregated_results} shows model accuracy on the QA task.
Overall, the proprietary models, GPT-5.4 and Claude Opus 4.6, outperform all open-weight models by at least 14 percentage points on the templated variant.
Because Qwen3 4B Instruct is used in the simulator, one might expect it to perform well, but this is not the case.
The difference between proprietary and open-weight models is statistically significant ($p\leq 0.001$) for both the templated and GPT-OSS 20B paraphrased variants.

Figure~\ref{fig:accuracy_by_question_category} shows the breakdown by question category.
All models perform well on the Repository Metadata category, confirming that they are capable of accessing data in a README file.
The next-best-performing category is Bivariate Statistics, likely because its ``hypothesis'' questions are three-class problems (yes/no/not possible), making them easier than questions with continuous answers.
The largest inter-model gap appears in the File Metadata, Directory Traversal, and Univariate Statistics categories, driven by the ``File Metadata-Count Rows'', ``Directory Traversal-Condition'', and ``Univariate Statistics-Condition'' question types.
These are the hardest questions: answering them correctly requires mapping question conditions onto the path and file schema, writing filtering code, and finally either computing the answer or determining that the question is unanswerable.
Appendix~\ref{sec:example_response} shows an example of a ``File Metadata-Count Rows'' question and how different LLMs apply different strategies when answering it.

To better understand how models fail, we manually annotate 100 incorrect QA pairs per model, paraphrased using GPT-OSS 20B (our best paraphraser, Appendix~\ref{sec:paraphrase_validity}).
The error categories (navigation, schema grounding, filtering, statistic computation, abstention, and answer formatting) are sequential.
Reaching filtering, for example, requires successful navigation and schema grounding.
Accordingly, each QA pair is annotated with its first encountered error.
Figure~\ref{fig:errors} shows that all models struggle with filtering, which involves selecting both files within directories and rows within files.
The higher-performing GPT-5.4 and Claude Opus 4.6 struggle less with navigation and schema grounding than the lower-performing models, suggesting that they progress further before making an error.
Because errors are sequential, low abstention-error rates do not imply strong abstention behavior.
Rather, models that fail earlier never reach a meaningful abstention decision.

\begin{figure}[t]
    \centering
    \vspace{-1em}
    \includegraphics[width=0.5\linewidth]{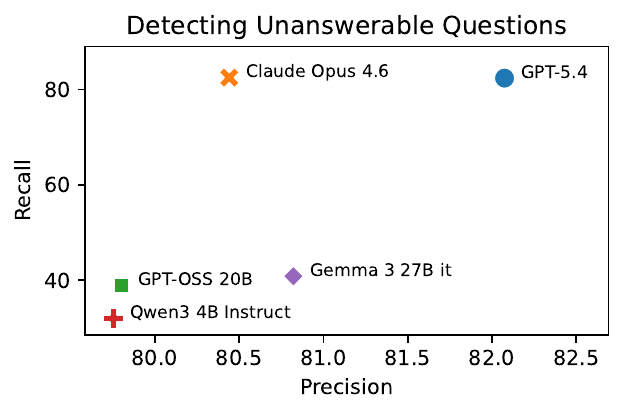}
    \caption{Each evaluated model's precision and recall for detecting unanswerable questions.}
    \label{fig:unanswerable}
    \vspace{-0.5em}
\end{figure}

\paragraph{Responding appropriately to unanswerable questions.}
Figure~\ref{fig:unanswerable} shows precision and recall for detecting unanswerable questions.
The proprietary models again lead, with both Claude Opus 4.6 and GPT-5.4 scoring above 80\% on both metrics, though neither exceeds 83\%. 
Open-weight models instead show high precision and low recall. 
They tend to answer when no answer exists, but are often correct when they abstain.
Taken with the error types in Figure~\ref{fig:errors}, these models' abstention capabilities are difficult to interpret, since they rarely progress deep enough into a problem for abstention to be relevant.

\begin{figure}[t]
    \centering
    \vspace{-1em}
    \includegraphics[width=\linewidth]{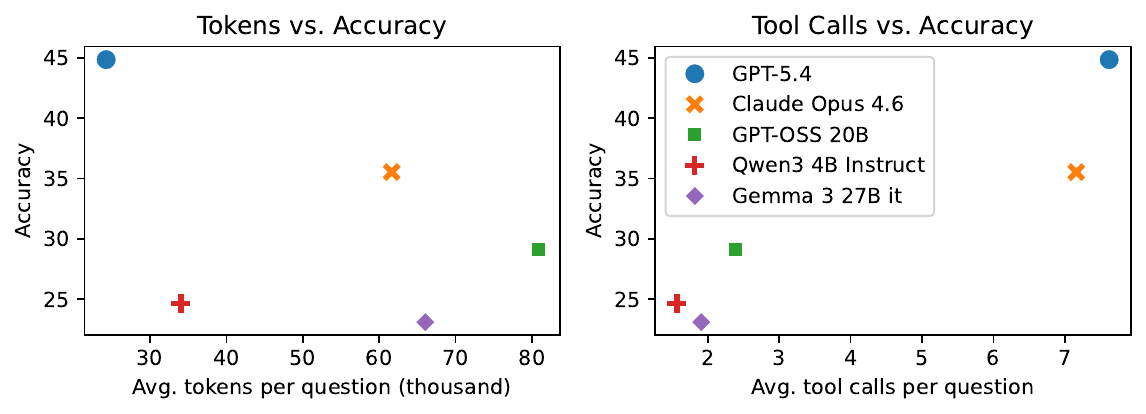}
    \caption{Comparisons between accuracy, token use, and tool use, per model. Although increased tool use correlates with accuracy, token use does not because high token use models typically attempt to load files into the LLM's context window, a suboptimal strategy.}
    \label{fig:accuracy_and_efficiency}
    \vspace{-0.5em}
\end{figure}

\paragraph{Programmatic analysis beats context stuffing.}
Figure~\ref{fig:accuracy_and_efficiency} breaks down performance by total tokens used and the number of tool calls made, aggregated over the templated questions and all paraphrases.
Although prior work shows that allocating additional test-time compute can improve performance~\citep{snell2025scaling}, higher-accuracy models in our evaluation do not use more tokens than lower-accuracy models.
In fact, GPT-5.4 achieves the highest accuracy while using the fewest tokens, averaging about 24,000 tokens per question.
The tool-call plot shows the opposite pattern: greater tool use is associated with higher accuracy.

Together, these figures suggest that higher-performing models use the available context window and tools more efficiently.
While lower-accuracy models tend to load large quantities of data into the context window, higher-performing models more often use the code interpreter to compute the relevant quantities.
An example of this is shown in Appendix~\ref{sec:example_response}.

\paragraph{Does paraphrasing introduce noise?}
Table~\ref{tab:paraphrase_agreement} shows Krippendorff's alpha~\citep{krippendorff2004content} for agreement between model responses to templated and paraphrased questions.
Scores for the three paraphrasing models range from 0.68 to 0.71, indicating moderate agreement.
Among the three paraphrasing models, GPT-OSS 20B has the highest score, indicating that its paraphrases elicit model responses most consistent with those elicited by the templated questions.
Interestingly, models generally score higher on their own paraphrases than those generated by a different model. 
This is consistent with models being better able to resolve ambiguities introduced by their own paraphrases.

To assess paraphrase quality directly, we audit 50 randomly sampled questions.
We manually inspect the paraphrased and templated variants of the questions, evaluating semantic equivalence, introduced ambiguity, and fluency.
GPT-OSS 20B scores the highest, with its paraphrases deemed 96\% semantically equivalent, 98\% unambiguous, and 100\% fluent.
The full results are available in Appendix~\ref{sec:paraphrase_validity}.
\section{\InfiniteScienceGym as a Unit Test and a Robustness Probe}
\label{sec:discussion}

\InfiniteScienceGym evaluates a narrow but important aspect of scientific assistance: whether a model can inspect a data repository, identify relevant evidence, compute a requested quantity, and recognize when the available data is insufficient.
Across our experiments, current models struggle to execute this sequence reliably.

Our results suggest three main takeaways.
First (\textbf{RQ1}), the proprietary models we evaluate outperform the open-weight models, but no model performs particularly well.
Repository-grounded reasoning requires models to combine navigation, schema grounding, filtering, and computation correctly, and errors at any stage can prevent successful completion.

Second (\textbf{RQ2}), no model handles unanswerable questions reliably in all cases, and open-weight models in particular miss many unanswerable cases.
This matters because, in realistic scientific settings, recognizing when the available evidence does not support a conclusion is itself an essential capability.
However, abstention performance must be interpreted together with earlier-stage errors. 
Models that fail early on may never reach the point at which deciding whether to abstain becomes meaningful.

Third (\textbf{RQ3}), higher accuracy is associated with more tool use, but not higher token usage.
Higher-performing models gather evidence selectively and operate programmatically, while lower-performing models load more data into the context window and reason from there.
Final-answer accuracy alone misses an important part of model behavior because scientific assistants must not only reach correct conclusions, but do so in the right way.

These results also clarify what kind of benchmark \InfiniteScienceGym is.
Each question targets a specific capability (e.g., traversal, filtering, aggregation, or abstention), making the benchmark a collection of \textbf{unit tests} for repository-grounded scientific reasoning.
It also functions as a \textbf{robustness benchmark}, since models must handle variation in repository structure and question wording.
Finally, it has aspects of a \textbf{counterfactual benchmark}.
Because the simulator controls which variables and relationships exist in a repository, it can construct cases, particularly unanswerable ones, that are difficult to isolate reliably in benchmarks derived from published studies.
We do not see these as competing interpretations, but as complementary perspectives on the same evaluation framework.

\InfiniteScienceGym's main advantage is control rather than realism. 
It is not a replacement for benchmarks grounded in real scientific artifacts, but a complement to real-data evaluations that is useful for stress-testing answerability, abstention, and data analysis under conditions where ground truth is known exactly.
\section{Conclusion}
\label{sec:conclusion}

In this paper, we introduced \InfiniteScienceGym, a simulator that generates realistic scientific repositories on demand, verifiable answerable and unanswerable questions, and naturalistic paraphrases of those questions.
The result is a controlled, storage-free benchmark that complements real-data evaluations by directly targeting abstention and evidence-grounded reasoning. 
Our experiments reveal that no evaluated model exceeds 50\% accuracy, that unanswerable questions remain a major source of failure, and that higher-performing models tend to use tools more effectively rather than simply consuming more tokens.

\begin{table}[t]
    \vspace{-1em}
    \centering
    \begin{tabular}{l|rrr}
        \toprule
         & \multicolumn{3}{c}{\textbf{Paraphrase Model}} \\
       \textbf{Evaluated Model} & GPT-OSS 20B & Qwen3 4B Inst. & Gemma 3 27B it \\
        \midrule
        GPT-5.4 & 0.74 & 0.66 & 0.58 \\
        Claude Opus 4.6 & 0.59 & 0.49 & 0.48 \\
        GPT-OSS 20B & \cellcolor{red!15} 0.72 & 0.69 & 0.67 \\
        Qwen3 4B Inst. & 0.76 & \cellcolor{red!15} 0.80 & 0.83 \\
        Gemma 3 27B it & 0.81 & 0.83 & \cellcolor{red!15} 0.85 \\
        \midrule
        Average & 0.72 & 0.69 & 0.68 \\
        \bottomrule        
    \end{tabular}
    \caption{Agreement between per-question correctness on templated and paraphrased variants, measured using Krippendorff's alpha. Rows denote evaluated models and columns denote paraphrasing models. Red cells indicate cases where the evaluated model and paraphrasing model are the same.}
    \label{tab:paraphrase_agreement}
    \vspace{-0.5em}
\end{table}

A distinctive feature is that unanswerable questions are guaranteed by construction, rather than identified through expert annotation or uncertain automated filters. 
This allows us to directly measure abstention in a way that real-data benchmarks cannot easily support.

\InfiniteScienceGym is intended primarily as an evaluation framework.
Strong performance on it should not be taken as evidence of broad scientific competence.
The benchmark focueses empirical tabular data, leaving out images, video, audio, or non-empirical scientific reasoning.
Procedural generation may also introduce regularities that future models could exploit.
Finally, our notion of unanswerability is operational and benchmark-specific, and narrower than the uncertainty encountered in real scientific practice.

Several future directions follow naturally.
The simulator could be extended to additional data modalities and messier repository structures to increase realism.
Control over the data-generating process could also be used to construct repositories that contradict plausible prior knowledge, enabling targeted measurement of known-knowledge bias.
Privileged access to the simulator opens a novel way to study hallucination by checking whether the data required to answer a question is ever accessed by the LLM.
The unbounded generation process also creates opportunities for model training. 
However, training on simulated data would require careful external evaluation to ensure that models do not overfit simulator-specific regularities at the expense of broader scientific-analysis capabilities.
More broadly, the controlled and infinite nature of \InfiniteScienceGym opens opportunities for studying how scientific assistants acquire, verify, and use evidence.

\section*{Reproducibility Statement}
\label{sec:reproducibility_statement}

Following the ACM definitions of repeatability, reproducibility, and replicability~\citep{acmartifactbadging}, \InfiniteScienceGym is designed to maximize \textbf{repeatability} and \textbf{reproducibility} through seed-driven generation.
Given the same seed, the simulator deterministically generates the same repository, file contents, and question-answer pairs.
Thus, the benchmark instances and ground-truth labels used in this work can be regenerated exactly from the released code and seeds.

The main exception is model evaluation.
Our reported results depend on sampled model outputs, which are not always deterministic.
For open-weight models, this variation can largely be controlled by fixing decoding settings and random seeds.
For proprietary models, exact evaluation reproducibility is more limited because sampling behavior and backend changes are not fully under the user's control.
We therefore claim exact reproducibility for the benchmark artifacts, but only partial reproducibility for the evaluation results.
We do not claim replicability in the ACM sense, though we hope the released code, seeds, and evaluation setup make it possible.

It should also be noted that exact byte-for-byte regeneration depends on library versions and hardware.
To address this, the accompanying code explicitly documents the pinned library versions, as well as the exact GPU hardware specifications.
\section*{Acknowledgements}

We are grateful to the reviewers for their insightful discussions and constructive criticism.
We thank the members of the Utah NLP lab, especially prof. Kenneth Marino and JK Refsgaard Schou, for their help defining our problem, planning and interpreting our experiments, annotating in our user study, and providing input on our manuscript.
We also thank prof. Jacob Hochhalter and Daisy Quach for their feedback about how the simulator fits into specific \emph{AI-for-science} use cases.
We are grateful to all the participants in the human study who provided annotations.
We also thank the Center for High Performance Computing at the University of Utah for their support and GPU resources, without which we would not be able to perform our experiments.
This material is based upon work supported by the National Science Foundation under Grant Numbers 2411319 and 2217154.
Any opinions, findings, and conclusions or recommendations expressed in this material are those of the author(s) and do not necessarily reflect the views of the National Science Foundation.

\bibliography{InfiniteScienceGym}

@inproceedings{
majumder2025discoverybench,
title={DiscoveryBench: Towards Data-Driven Discovery with Large Language Models},
author={Bodhisattwa Prasad Majumder and Harshit Surana and Dhruv Agarwal and Bhavana Dalvi Mishra and Abhijeetsingh Meena and Aryan Prakhar and Tirth Vora and Tushar Khot and Ashish Sabharwal and Peter Clark},
booktitle={The Thirteenth International Conference on Learning Representations},
year={2025},
url={https://openreview.net/forum?id=vyflgpwfJW}
}

@inproceedings{
chen2025scienceagentbench,
title={ScienceAgentBench: Toward Rigorous Assessment of Language Agents for Data-Driven Scientific Discovery},
author={Ziru Chen and Shijie Chen and Yuting Ning and Qianheng Zhang and Boshi Wang and Botao Yu and Yifei Li and Zeyi Liao and Chen Wei and Zitong Lu and Vishal Dey and Mingyi Xue and Frazier N. Baker and Benjamin Burns and Daniel Adu-Ampratwum and Xuhui Huang and Xia Ning and Song Gao and Yu Su and Huan Sun},
booktitle={The Thirteenth International Conference on Learning Representations},
year={2025},
url={https://openreview.net/forum?id=6z4YKr0GK6}
}

@inproceedings{yao2023react,
  title = {{ReAct}: Synergizing Reasoning and Acting in Language Models},
  author = {Yao, Shunyu and Zhao, Jeffrey and Yu, Dian and Du, Nan and Shafran, Izhak and Narasimhan, Karthik and Cao, Yuan},
  booktitle = {International Conference on Learning Representations (ICLR) },
  year = {2023},
  html = {https://arxiv.org/abs/2210.03629},
}

@inproceedings{
shojaee2025llmsrbench,
title={{LLM}-{SRB}ench: A New Benchmark for Scientific Equation Discovery with Large Language Models},
author={Parshin Shojaee and Ngoc-Hieu Nguyen and Kazem Meidani and Amir Barati Farimani and Khoa D Doan and Chandan K. Reddy},
booktitle={Forty-second International Conference on Machine Learning},
year={2025},
url={https://openreview.net/forum?id=SyQPiZJVWY}
}

@inproceedings{
agarwal2025autodiscovery,
title={AutoDiscovery: Open-ended Scientific Discovery via Bayesian Surprise},
author={Dhruv Agarwal and Bodhisattwa Prasad Majumder and Reece Adamson and Megha Chakravorty and Satvika Reddy Gavireddy and Aditya Parashar and Harshit Surana and Bhavana Dalvi Mishra and Andrew McCallum and Ashish Sabharwal and Peter Clark},
booktitle={The Thirty-ninth Annual Conference on Neural Information Processing Systems},
year={2025},
url={https://openreview.net/forum?id=kJqTkj2HhF}
}

@inproceedings{
tian2024scicode,
title={SciCode: A Research Coding Benchmark Curated by Scientists},
author={Minyang Tian and Luyu Gao and Dylan Zhang and Xinan Chen and Cunwei Fan and Xuefei Guo and Roland Haas and Pan Ji and Kittithat Krongchon and Yao Li and Shengyan Liu and Di Luo and Yutao Ma and HAO TONG and Kha Trinh and Chenyu Tian and Zihan Wang and Bohao Wu and Shengzhu Yin and Minhui Zhu and Kilian Lieret and Yanxin Lu and Genglin Liu and Yufeng Du and Tianhua Tao and Ofir Press and Jamie Callan and Eliu A Huerta and Hao Peng},
booktitle={The Thirty-eight Conference on Neural Information Processing Systems Datasets and Benchmarks Track},
year={2024},
url={https://openreview.net/forum?id=ADLaALtdoG}
}

@article{corebench,
title = "CORE-Bench: Fostering the Credibility of Published Research Through a Computational Reproducibility Agent Benchmark",
author = "Siegel, \{Zachary S.\} and Sayash Kapoor and Nitya Nadgir and Benedikt Stroebl and Arvind Narayanan",
note = "Publisher Copyright: {\textcopyright} 2025, Transactions on Machine Learning Research. All rights reserved.",
year = "2025",
month = jan,
language = "American English",
volume = "2025-January",
pages = "1--31",
journal = "Transactions on Machine Learning Research",
issn = "2835-8856",
publisher = "Transactions on Machine Learning Research",

}

@misc{novikov2025alphaevolvecodingagentscientific,
      title={AlphaEvolve: A coding agent for scientific and algorithmic discovery}, 
      author={Alexander Novikov and Ngân Vũ and Marvin Eisenberger and Emilien Dupont and Po-Sen Huang and Adam Zsolt Wagner and Sergey Shirobokov and Borislav Kozlovskii and Francisco J. R. Ruiz and Abbas Mehrabian and M. Pawan Kumar and Abigail See and Swarat Chaudhuri and George Holland and Alex Davies and Sebastian Nowozin and Pushmeet Kohli and Matej Balog},
      year={2025},
      eprint={2506.13131},
      archivePrefix={arXiv},
      primaryClass={cs.AI},
      url={https://arxiv.org/abs/2506.13131}, 
}

@misc{guevara2026singleminusgluontreeamplitudes,
      title={Single-minus gluon tree amplitudes are nonzero}, 
      author={Alfredo Guevara and Alexandru Lupsasca and David Skinner and Andrew Strominger and Kevin Weil},
      year={2026},
      eprint={2602.12176},
      archivePrefix={arXiv},
      primaryClass={hep-th},
      url={https://arxiv.org/abs/2602.12176}, 
}

@misc{feng2026autonomousmathematicsresearch,
      title={Towards Autonomous Mathematics Research}, 
      author={Tony Feng and Trieu H. Trinh and Garrett Bingham and Dawsen Hwang and Yuri Chervonyi and Junehyuk Jung and Joonkyung Lee and Carlo Pagano and Sang-hyun Kim and Federico Pasqualotto and Sergei Gukov and Jonathan N. Lee and Junsu Kim and Kaiying Hou and Golnaz Ghiasi and Yi Tay and YaGuang Li and Chenkai Kuang and Yuan Liu and Hanzhao Lin and Evan Zheran Liu and Nigamaa Nayakanti and Xiaomeng Yang and Heng-Tze Cheng and Demis Hassabis and Koray Kavukcuoglu and Quoc V. Le and Thang Luong},
      year={2026},
      eprint={2602.10177},
      archivePrefix={arXiv},
      primaryClass={cs.LG},
      url={https://arxiv.org/abs/2602.10177}, 
}

@inproceedings{tang-etal-2025-large,
    title = "Large Language Models for Automated Literature Review: An Evaluation of Reference Generation, Abstract Writing, and Review Composition",
    author = "Tang, Xuemei  and
      Duan, Xufeng  and
      Cai, Zhenguang",
    editor = "Christodoulopoulos, Christos  and
      Chakraborty, Tanmoy  and
      Rose, Carolyn  and
      Peng, Violet",
    booktitle = "Proceedings of the 2025 Conference on Empirical Methods in Natural Language Processing",
    month = nov,
    year = "2025",
    address = "Suzhou, China",
    publisher = "Association for Computational Linguistics",
    url = "https://aclanthology.org/2025.emnlp-main.83/",
    doi = "10.18653/v1/2025.emnlp-main.83",
    pages = "1602--1617",
    ISBN = "979-8-89176-332-6"
}

@misc{kalai2025languagemodelshallucinate,
      title={Why Language Models Hallucinate}, 
      author={Adam Tauman Kalai and Ofir Nachum and Santosh S. Vempala and Edwin Zhang},
      year={2025},
      eprint={2509.04664},
      archivePrefix={arXiv},
      primaryClass={cs.CL},
      url={https://arxiv.org/abs/2509.04664}, 
}

@inproceedings{xia-etal-2025-survey,
    title = "A Survey of Uncertainty Estimation Methods on Large Language Models",
    author = "Xia, Zhiqiu  and
      Xu, Jinxuan  and
      Zhang, Yuqian  and
      Liu, Hang",
    editor = "Che, Wanxiang  and
      Nabende, Joyce  and
      Shutova, Ekaterina  and
      Pilehvar, Mohammad Taher",
    booktitle = "Findings of the Association for Computational Linguistics: ACL 2025",
    month = jul,
    year = "2025",
    address = "Vienna, Austria",
    publisher = "Association for Computational Linguistics",
    url = "https://aclanthology.org/2025.findings-acl.1101/",
    doi = "10.18653/v1/2025.findings-acl.1101",
    pages = "21381--21396",
    ISBN = "979-8-89176-256-5"
}

@inproceedings{jia-etal-2025-simulbench,
    title = "{S}imul{B}ench: Evaluating Language Models with Creative Simulation Tasks",
    author = "Jia, Qi  and
      Yue, Xiang  and
      Zheng, Tuney  and
      Huang, Jie  and
      Lin, Bill Yuchen",
    editor = "Chiruzzo, Luis  and
      Ritter, Alan  and
      Wang, Lu",
    booktitle = "Findings of the Association for Computational Linguistics: NAACL 2025",
    month = apr,
    year = "2025",
    address = "Albuquerque, New Mexico",
    publisher = "Association for Computational Linguistics",
    url = "https://aclanthology.org/2025.findings-naacl.453/",
    doi = "10.18653/v1/2025.findings-naacl.453",
    pages = "8133--8146",
    ISBN = "979-8-89176-195-7"
}

@inproceedings{
yin2023alcuna,
title={{ALCUNA}: Large Language Models Meet New Knowledge},
author={Xunjian Yin and Baizhou Huang and Xiaojun Wan},
booktitle={The 2023 Conference on Empirical Methods in Natural Language Processing},
year={2023},
url={https://openreview.net/forum?id=toUPGCAMic}
}

@inproceedings{feng-etal-2024-dont,
    title = "Don{'}t Hallucinate, Abstain: Identifying {LLM} Knowledge Gaps via Multi-{LLM} Collaboration",
    author = "Feng, Shangbin  and
      Shi, Weijia  and
      Wang, Yike  and
      Ding, Wenxuan  and
      Balachandran, Vidhisha  and
      Tsvetkov, Yulia",
    editor = "Ku, Lun-Wei  and
      Martins, Andre  and
      Srikumar, Vivek",
    booktitle = "Proceedings of the 62nd Annual Meeting of the Association for Computational Linguistics (Volume 1: Long Papers)",
    month = aug,
    year = "2024",
    address = "Bangkok, Thailand",
    publisher = "Association for Computational Linguistics",
    url = "https://aclanthology.org/2024.acl-long.786/",
    doi = "10.18653/v1/2024.acl-long.786",
    pages = "14664--14690"
}

@article{wen-etal-2025-know,
    title = "Know Your Limits: A Survey of Abstention in Large Language Models",
    author = "Wen, Bingbing  and
      Yao, Jihan  and
      Feng, Shangbin  and
      Xu, Chenjun  and
      Tsvetkov, Yulia  and
      Howe, Bill  and
      Wang, Lucy Lu",
    journal = "Transactions of the Association for Computational Linguistics",
    volume = "13",
    year = "2025",
    address = "Cambridge, MA",
    publisher = "MIT Press",
    url = "https://aclanthology.org/2025.tacl-1.26/",
    doi = "10.1162/tacl_a_00754",
    pages = "529--556"
}

@ARTICLE{Dickersin1993-ij,
  title    = "Publication bias: the problem that won't go away",
  author   = "Dickersin, K and Min, Y I",
  journal  = "Ann N Y Acad Sci",
  volume   =  703,
  pages    = "135--46; discussion 146--8",
  month    =  dec,
  year     =  1993,
  address  = "United States",
  language = "en"
}

@ARTICLE{Nissen2016-fe,
  title    = "Publication bias and the canonization of false facts",
  author   = "Nissen, Silas Boye and Magidson, Tali and Gross, Kevin and
              Bergstrom, Carl T",
  journal  = "Elife",
  volume   =  5,
  month    =  dec,
  year     =  2016,
  address  = "England",
  language = "en"
}

@misc{acmartifactbadging,
  author       = {{Association for Computing Machinery}},
  title        = {Artifact Review and Badging},
  year         = {2024},
  howpublished = {\url{https://www.acm.org/publications/policies/artifact-review-and-badging-current}},
  note         = {Accessed: 2026-03-29}
}

@inproceedings{gupta-etal-2020-infotabs,
    title = "{INFOTABS}: Inference on Tables as Semi-structured Data",
    author = "Gupta, Vivek  and
      Mehta, Maitrey  and
      Nokhiz, Pegah  and
      Srikumar, Vivek",
    editor = "Jurafsky, Dan  and
      Chai, Joyce  and
      Schluter, Natalie  and
      Tetreault, Joel",
    booktitle = "Proceedings of the 58th Annual Meeting of the Association for Computational Linguistics",
    month = jul,
    year = "2020",
    address = "Online",
    publisher = "Association for Computational Linguistics",
    url = "https://aclanthology.org/2020.acl-main.210/",
    doi = "10.18653/v1/2020.acl-main.210",
    pages = "2309--2324"
}

@inproceedings{TableBench,
author = {Wu, Xianjie and Yang, Jian and Chai, Linzheng and Zhang, Ge and Liu, Jiaheng and Du, Xeron and Liang, Di and Shu, Daixin and Cheng, Xianfu and Sun, Tianzhen and Li, Tongliang and Li, Zhoujun and Niu, Guanglin},
title = {TableBench: a comprehensive and complex benchmark for table question answering},
year = {2025},
isbn = {978-1-57735-897-8},
publisher = {AAAI Press},
url = {https://doi.org/10.1609/aaai.v39i24.34739},
doi = {10.1609/aaai.v39i24.34739},
booktitle = {Proceedings of the Thirty-Ninth AAAI Conference on Artificial Intelligence and Thirty-Seventh Conference on Innovative Applications of Artificial Intelligence and Fifteenth Symposium on Educational Advances in Artificial Intelligence},
articleno = {2841},
numpages = {10},
series = {AAAI'25/IAAI'25/EAAI'25}
}

@inproceedings{
Chen2020TabFact,
title={TabFact: A Large-scale Dataset for Table-based Fact Verification},
author={Wenhu Chen and Hongmin Wang and Jianshu Chen and Yunkai Zhang and Hong Wang and Shiyang Li and Xiyou Zhou and William Yang Wang},
booktitle={International Conference on Learning Representations},
year={2020},
url={https://openreview.net/forum?id=rkeJRhNYDH}
}

@inproceedings{
    jimenez2024swebench,
    title={{SWE}-bench: Can Language Models Resolve Real-world Github Issues?},
    author={Carlos E Jimenez and John Yang and Alexander Wettig and Shunyu Yao and Kexin Pei and Ofir Press and Karthik R Narasimhan},
    booktitle={The Twelfth International Conference on Learning Representations},
    year={2024},
    url={https://openreview.net/forum?id=VTF8yNQM66}
}

@misc{liu2023repobench,
      title={RepoBench: Benchmarking Repository-Level Code Auto-Completion Systems}, 
      author={Tianyang Liu and Canwen Xu and Julian McAuley},
      year={2024},
      url={https://arxiv.org/abs/2306.03091},
      booktitle={International Conference on Learning Representations}
}

@article{li2024can,
  title={Can llm already serve as a database interface? a big bench for large-scale database grounded text-to-sqls},
  author={Li, Jinyang and Hui, Binyuan and Qu, Ge and Yang, Jiaxi and Li, Binhua and Li, Bowen and Wang, Bailin and Qin, Bowen and Geng, Ruiying and Huo, Nan and others},
  journal={Advances in Neural Information Processing Systems},
  volume={36},
  year={2024}
}

@inproceedings{
lei2025spider,
title={Spider 2.0: Evaluating Language Models on Real-World Enterprise Text-to-{SQL} Workflows},
author={Fangyu Lei and Jixuan Chen and Yuxiao Ye and Ruisheng Cao and Dongchan Shin and Hongjin SU and ZHAOQING SUO and Hongcheng Gao and Wenjing Hu and Pengcheng Yin and Victor Zhong and Caiming Xiong and Ruoxi Sun and Qian Liu and Sida Wang and Tao Yu},
booktitle={The Thirteenth International Conference on Learning Representations},
year={2025},
url={https://openreview.net/forum?id=XmProj9cPs}
}

@inproceedings{
snell2025scaling,
title={Scaling {LLM} Test-Time Compute Optimally Can be More Effective than Scaling Parameters for Reasoning},
author={Charlie Victor Snell and Jaehoon Lee and Kelvin Xu and Aviral Kumar},
booktitle={The Thirteenth International Conference on Learning Representations},
year={2025},
url={https://openreview.net/forum?id=4FWAwZtd2n}
}

@Misc{smolagents,
  title =        {`smolagents`: a smol library to build great agentic systems.},
  author =       {Aymeric Roucher and Albert Villanova del Moral and Thomas Wolf and Leandro von Werra and Erik Kaunismäki},
  howpublished = {\url{https://github.com/huggingface/smolagents}},
  year =         {2025}
}

@inproceedings{troshin-etal-2025-asking,
    title = "Asking a Language Model for Diverse Responses",
    author = "Troshin, Sergey  and
      Saparina, Irina  and
      Fokkens, Antske  and
      Niculae, Vlad",
    editor = {Eikema, Bryan  and
      V{\'a}zquez, Ra{\'u}l  and
      Berant, Jonathan  and
      de Marneffe, Marie-Catherine  and
      Plank, Barbara  and
      Shelmanov, Artem  and
      Swayamdipta, Swabha  and
      Tiedemann, J{\"o}rg  and
      Zerva, Chrysoula  and
      Aziz, Wilker},
    booktitle = "Proceedings of the 2nd Workshop on Uncertainty-Aware NLP (UncertaiNLP 2025)",
    month = nov,
    year = "2025",
    address = "Suzhou, China",
    publisher = "Association for Computational Linguistics",
    url = "https://aclanthology.org/2025.uncertainlp-main.8/",
    doi = "10.18653/v1/2025.uncertainlp-main.8",
    pages = "66--72",
    ISBN = "979-8-89176-349-4"
}

@book{krippendorff2004content,
  title={Content analysis: An introduction to its methodology},
  author={Krippendorff, Klaus},
  year={2004},
  publisher={Sage},
  address={Thousand Oaks, CA},
  edition={2nd}
}

@misc{qwen3technicalreport,
      title={Qwen3 Technical Report}, 
      author={Qwen Team},
      year={2025},
      eprint={2505.09388},
      archivePrefix={arXiv},
      primaryClass={cs.CL},
      url={https://arxiv.org/abs/2505.09388}, 
}

@misc{gemma_2025,
      title={Gemma 3 Technical Report}, 
      author={Gemma Team},
      year={2025},
      eprint={2503.19786},
      archivePrefix={arXiv},
      primaryClass={cs.CL},
      url={https://arxiv.org/abs/2503.19786},
}

@misc{openai2025gptoss120bgptoss20bmodel,
      title={gpt-oss-120b \& gpt-oss-20b Model Card}, 
      author={OpenAI and : and Sandhini Agarwal and Lama Ahmad and Jason Ai and Sam Altman and Andy Applebaum and Edwin Arbus and Rahul K. Arora and Yu Bai and Bowen Baker and Haiming Bao and Boaz Barak and Ally Bennett and Tyler Bertao and Nivedita Brett and Eugene Brevdo and Greg Brockman and Sebastien Bubeck and Che Chang and Kai Chen and Mark Chen and Enoch Cheung and Aidan Clark and Dan Cook and Marat Dukhan and Casey Dvorak and Kevin Fives and Vlad Fomenko and Timur Garipov and Kristian Georgiev and Mia Glaese and Tarun Gogineni and Adam Goucher and Lukas Gross and Katia Gil Guzman and John Hallman and Jackie Hehir and Johannes Heidecke and Alec Helyar and Haitang Hu and Romain Huet and Jacob Huh and Saachi Jain and Zach Johnson and Chris Koch and Irina Kofman and Dominik Kundel and Jason Kwon and Volodymyr Kyrylov and Elaine Ya Le and Guillaume Leclerc and James Park Lennon and Scott Lessans and Mario Lezcano-Casado and Yuanzhi Li and Zhuohan Li and Ji Lin and Jordan Liss and Lily and Liu and Jiancheng Liu and Kevin Lu and Chris Lu and Zoran Martinovic and Lindsay McCallum and Josh McGrath and Scott McKinney and Aidan McLaughlin and Song Mei and Steve Mostovoy and Tong Mu and Gideon Myles and Alexander Neitz and Alex Nichol and Jakub Pachocki and Alex Paino and Dana Palmie and Ashley Pantuliano and Giambattista Parascandolo and Jongsoo Park and Leher Pathak and Carolina Paz and Ludovic Peran and Dmitry Pimenov and Michelle Pokrass and Elizabeth Proehl and Huida Qiu and Gaby Raila and Filippo Raso and Hongyu Ren and Kimmy Richardson and David Robinson and Bob Rotsted and Hadi Salman and Suvansh Sanjeev and Max Schwarzer and D. Sculley and Harshit Sikchi and Kendal Simon and Karan Singhal and Yang Song and Dane Stuckey and Zhiqing Sun and Philippe Tillet and Sam Toizer and Foivos Tsimpourlas and Nikhil Vyas and Eric Wallace and Xin Wang and Miles Wang and Olivia Watkins and Kevin Weil and Amy Wendling and Kevin Whinnery and Cedric Whitney and Hannah Wong and Lin Yang and Yu Yang and Michihiro Yasunaga and Kristen Ying and Wojciech Zaremba and Wenting Zhan and Cyril Zhang and Brian Zhang and Eddie Zhang and Shengjia Zhao},
      year={2025},
      eprint={2508.10925},
      archivePrefix={arXiv},
      primaryClass={cs.CL},
      url={https://arxiv.org/abs/2508.10925}, 
}
\bibliographystyle{colm2026_conference}

\appendix

\clearpage

\section{Scientific Taxonomy}
\label{sec:scientific_taxonomy}

\begin{table}[h]
    \centering
    \begin{tabular}{l|r|r}
        \toprule
        Field & \# Domains & \# Subdomains \\
        \midrule
        Computer Science & 11 & 41 \\
        Engineering - Bioengineering & 9 & 32 \\
        Engineering - Chemical and Materials & 7 & 31 \\
        Engineering - Civil and Environmental & 9 & 25 \\
        Engineering - Electrical and Computer & 17 & 59 \\
        Engineering - Industrial and Systems & 8 & 24 \\
        Engineering - Mechanical and Aerospace & 9 & 29 \\
        Life Science - Agricultural and Plant & 19 & 49 \\
        Life Science - Biology & 9 & 45 \\
        Life Science - Biomedical and Health & 25 & 68 \\
        Life Science - Neuroscience & 17 & 48 \\
        Natural - Astronomy and Space & 6 & 34 \\
        Natural - Chemistry & 12 & 43 \\
        Natural - Earth and Environmental & 18 & 59 \\
        Natural - Physics & 18 & 47 \\
        Social - Economics & 8 & 18 \\
        Social - Education & 9 & 21 \\
        Social - Geography and Demography & 6 & 21 \\
        Social - Political Science & 6 & 20 \\
        Social - Psychology and Cognitive & 9 & 27 \\
        Social - Sociology and Anthropology & 9 & 21 \\
        Math - Statistics and Applied & 3 & 18 \\
        \bottomrule
    \end{tabular}
    \caption{The numbers of domains and subdomains per field in the scientific taxonomy}
    \label{tab:taxonomy}
\end{table}

The taxonomy is broken down into fields, domains, and subdomains.
When creating a scientific repository, we first sample a field, then one of the field's domains, and finally one of the domain's subdomains.
Table~\ref{tab:taxonomy} shows a breakdown of the fields used in our taxonomy, along with the numbers of domains and subdomains per field.

The scientific taxonomy is not intended to be an exceptionally accurate representation of existing scientific areas.
Rather, its purpose is to capture \emph{most} areas of active, empirical science and introduce diversity in the simulated data.

\section{Example Project Metadata}
\label{sec:full_example}

The following subsections show the project description, abstract, path variables, and file variables corresponding to repository \#118.

\subsection{Repository \#118 Description}

This research project, titled 'Omics-Driven Strain Optimization for Ethanol', is a real-world, empirically grounded study conducted in a biotechnology research lab focused on industrial biofuel production. The goal is to identify and engineer microbial strains—specifically Saccharomyces cerevisiae—that produce higher ethanol yields under varying fermentation conditions by leveraging multi-omics data and high-throughput phenotyping. The project integrates genomics, transcriptomics, metabolomics, and proteomics to understand the molecular mechanisms underlying ethanol productivity, with the ultimate aim of optimizing strain performance through data-driven selection.

The research question is: Which combinations of genetic modifications and environmental conditions lead to the highest ethanol yield per unit of glucose consumed, while maintaining strain stability and metabolic robustness?

We test a panel of 12 genetically modified S. cerevisiae strains (each with targeted knockouts or overexpressions of genes involved in glycolysis, ethanol metabolism, and stress response), grown under 6 different fermentation conditions (varying pH, temperature, glucose concentration, and oxygen levels). Each strain is grown in triplicate across conditions, and we measure multiple phenotypic and omics outputs. The study is designed to be scalable and reproducible, with all data encoded as flat, tabular files (CSV or TSV) to ensure compatibility with downstream statistical and machine learning analyses.

Independent variables (experimental factors):
\begin{itemize}[nosep]
    \item Strain ID (e.g., WT, $\Delta$pfk1, $\Delta$pyk1, pYKR-ADH2, etc.)
    \item Fermentation temperature (15°C, 25°C, 30°C, 35°C)
    \item pH (4.0, 5.0, 6.0, 7.0)
    \item Glucose concentration (50 g/L, 100 g/L, 200 g/L)
    \item Oxygen level (anaerobic, microaerobic, aerated)
    \item Growth phase (early, mid, late)
    \item Time of ethanol measurement (0, 6, 12, 24, 48 hours)
    \item Genotype type (knockout, overexpression, promoter swap)
\end{itemize}

Dependent variables (measured outcomes):

\begin{itemize}[nosep]
    \item Ethanol yield (g/L)
    \item Glucose consumption rate (g/L/h)
    \item Biomass concentration (g/L)
    \item Residual glucose (g/L)
    \item Cell viability (\%)
    \item Ethanol purity (\%)
    \item Total metabolite concentration (e.g., glycerol, acetate, lactate) — measured via GC-MS
    \item Relative gene expression (RNA-seq normalized counts, log2-transformed)
    \item Protein abundance (LC-MS quantified, normalized to reference)
    \item Metabolic flux ratio (e.g., glycolysis vs. TCA cycle)
    \item Stress response markers (e.g., Hsp70, Msn2 expression)
    \item Fermentation stability index (defined as coefficient of variation of ethanol yield across replicates)
\end{itemize}

Potential confounders:

\begin{itemize}[nosep]
    \item Batch-to-batch variation in media composition
    \item Microbial contamination (detected via 16S rRNA or ITS sequencing)
    \item Inconsistent inoculum size or age
    \item Temperature fluctuations in incubators
    \item Differences in gas exchange or CO2 scrubbing efficiency
    \item Variability in GC-MS calibration or quantification
\end{itemize}

Each data file represents a single experimental run (one strain × one condition × one time point). For example, a file named 'strain\_$\Delta$pyk1\_pH5.0\_temp30C\_glucose100g\_24h.csv' contains one row per replicate (triplicate) of that condition. Each row includes all the independent and dependent variables listed above, with metadata such as strain ID, condition, time point, replicate number, and date/time of measurement. The data structure is designed for clarity: columns represent variables (e.g., 'ethanol\_yield', 'glucose\_consumption', 'gene\_expression\_pfkl'), and rows represent individual experimental observations.

Files are organized by strain and condition, with subdirectories for each strain (e.g.,'strain\_$\Delta$pfk1/') containing condition-specific folders (e.g., 'pH4.0/', 'pH5.0/') and within those, time-point-specific files. This structure allows for easy tracing of experimental design and facilitates multi-level analysis (e.g., strain-level vs. condition-level effects).

The significance of this project lies in its practical application to industrial biotechnology. Ethanol production is a cornerstone of renewable energy, and strain optimization directly impacts cost, efficiency, and scalability. Unlike traditional trial-and-error approaches, this project uses omics data to build predictive models that link molecular phenotypes to fermentation performance. This enables a move from empirical screening to intelligent, data-driven strain design. The novelty lies in integrating multi-omics data (transcriptome, proteome, metabolome) with high-throughput phenotypic measurements in a structured, reproducible, and scalable format—making it a model for future omics-based strain engineering in biofuel and bioprocess industries.

This project is not only scientifically rigorous but also methodologically transparent, with all data encoded in flat tables that can be easily shared, analyzed, and reused by other researchers.

\subsection{Repository \#118 Abstract}

This study presents an omics-driven approach to strain optimization for industrial ethanol production using genetically engineered Saccharomyces cerevisiae. By integrating genomics, transcriptomics, proteomics, and metabolomics with high-throughput phenotyping, we systematically evaluate how combinations of genetic modifications—such as knockouts and overexpressions in glycolysis and stress response pathways—interact with fermentation conditions (pH, temperature, glucose concentration, oxygen levels, and time) to influence ethanol yield and metabolic robustness. A panel of 12 strains is tested across 6 fermentation conditions in triplicate, enabling multi-level analysis of strain- and condition-specific performance. We quantify key phenotypes including ethanol yield, glucose consumption, biomass, and fermentation stability, while leveraging omics data to identify molecular signatures associated with high productivity. Through rigorous experimental design and controlled confounder management, we establish a scalable, reproducible framework for data-driven strain selection. This work advances synthetic biology by bridging molecular mechanisms with industrial performance, enabling predictive modeling of strain behavior under diverse bioprocess conditions. The open, flat-file data structure ensures transparency, reproducibility, and reusability, positioning this study as a model for future omics-guided strain engineering in biofuel and bioprocessing applications.

\subsection{Repository \#118 Path Variables}
\label{sec:118_path_variables}

\begin{table}[!htbp]
    \centering
    \makegapedcells
    \begin{tabular}{lll}
    \toprule
    \makecell[l]{Name (short name)} & Values & Description \\
    \midrule
    growth phase (gphase) & early, mid, late & \makecell[l]{The stage of microbial growth in the culture, \\ indicating whether cells are in early \\ (exponential), mid (mid-exponential), or late \\ (stationary) phase, with no physical units.} \\
    \makecell[l]{Genotype Modification \\ Type (gtype)} & \makecell[l]{knockout, \\ overexpression, \\ promoter\_swap} & \makecell[l]{The type of genetic modification applied to \\ the strain (e.g., gene knockout, \\ overexpression, or promoter swap), with no \\ physical units.} \\
    Date (date) & \makecell[l]{01/05/2025 to \\ 11/11/2025} & Date. Includes day, month and optional year. \\
    \makecell[l]{Time point of ethanol \\ measurement during \\ fermentation (tpt)} & 0, 6, 12, 24, 48 & \makecell[l]{The time at which ethanol yield and other \\ phenotypic measurements are taken during \\ fermentation, in hours.} \\
    \makecell[l]{Sequential Number \\ (seq\_number)} & 1, 2, 3, 4, 5 & A sequential trial or run number. \\
    fermentation pH (pH) & 4.0, 5.0, 6.0, 7.0 & \makecell[l]{The acidity or alkalinity of the fermentation \\ medium, measured on the pH scale (0–14), \\ with values ranging from 4.0 to 7.0.} \\
    \bottomrule
    \end{tabular}
    \caption{The variables used in the directory/file paths for repository \#118.}
    \label{tab:path_variables}
\end{table}

Table~\ref{tab:path_variables} shows the path variables included in repository \#118. 
The names, values, and descriptions are sampled from the LLM, conditioned on the project specification.

\subsection{Repository \#118 File Variables}
\label{sec:118_file_variables}

\begin{table}[!htbp]
    \centering
    \makegapedcells
    \begin{tabular}{llll}
    \toprule
    \makecell[l]{Name (short)} & Description & Type & Values (Probs.) \\
    \midrule
    \makecell[l]{Fermentation \\ temperature \\ (temp)} & \makecell[l]{The temperature at which \\ fermentation occurs, in \\ degrees Celsius} & ind/cat & \makecell[l]{15°C (25.0\%), \\ 25°C (25.0\%), \\ 30°C (25.0\%), \\ 35°C (25.0\%)} \\
    \makecell[l]{Glucose \\ concentration \\ (glucose\_conc)} & \makecell[l]{The initial concentration of \\ glucose in the fermentation \\ medium, in grams per liter} & ind/cat & \makecell[l]{50 g/L (33.3\%), \\ 100 g/L (33.3\%), \\ 200 g/L (33.3\%)} \\
    \makecell[l]{Oxygen level \\ (oxygen\_level)} & \makecell[l]{The level of oxygen present \\ during fermentation: \\ anaerobic, microaerobic, or \\ aerated} & ind/cat & \makecell[l]{anaerobic (50.0\%), \\ microaerobic (30.0\%), \\ aerated (20.0\%)} \\
    \makecell[l]{Ethanol yield \\ (ethanol\_yield)} & \makecell[l]{Amount of ethanol produced \\ per liter of fermentation \\ medium, in grams per liter} & dep/con &  \\
    \makecell[l]{Glucose \\ consumption \\ rate} & \makecell[l]{Rate at which glucose is \\ consumed, in grams per liter \\ per hour} & dep/con &  \\
    \makecell[l]{Biomass \\ concentration \\ (biomass)} & \makecell[l]{Total cell mass concentration, \\ in grams per liter} & dep/con &  \\
    \makecell[l]{Residual glucose \\ (residual\_glucose)} & \makecell[l]{Amount of glucose remaining \\ after fermentation, in grams \\ per liter} & dep/con &  \\
    \makecell[l]{Cell viability \\ (cell\_viability)} & \makecell[l]{Percentage of live cells in the \\ culture, measured via staining \\ or viability assay} & dep/con &  \\
    \makecell[l]{Ethanol purity \\ (ethanol\_purity)} & \makecell[l]{Percentage of ethanol in the \\ final fermentation mixture, \\ by volume} & dep/con &  \\
    \makecell[l]{Glycerol \\ concentration \\ (glycerol)} & \makecell[l]{Concentration of glycerol in \\ the fermentation broth, \\ measured via GC-MS, in \\ millimoles per liter} & dep/con &  \\
    \makecell[l]{Acetate \\ concentration \\ (acetate)} & \makecell[l]{Concentration of acetate in \\ the fermentation broth, \\ measured via GC-MS, in \\ millimoles per liter} & dep/con &  \\
    \makecell[l]{Fermentation \\ stability index \\ (stability\_index)} & \makecell[l]{A normalized measure of \\ ethanol yield consistency \\ across replicates, defined as \\ coefficient of variation} & dep/con & \\
    \bottomrule
    \end{tabular}
    \caption{The variables used as file columns for repository \#118. The first type value is either independent (ind) or dependent (dep), and the second value is either categorical (cat) or continuous (con). The last column shows the values and probabilities for independent, categorical variables. The short name for ``Glucose consumption rate'' (omitted for formatting) is ``glucose\_consumption\_rate''.}
    \label{tab:file_variables}
\end{table}

Table~\ref{tab:file_variables} shows the file variables for repository \#118.
These variables are used to populate the columns of every file.
While this repository only has independent, categorical variables and dependent, continuous variables, it could also contain independent, continuous variables and dependent, categorical variables.
This repository coincidentally does not contain those variable types.

Although we don't include the Python code for every dependent variable, Appendix~\ref{sec:dependent_variables} contains the code that defines the ``Glucose consumption rate'' variable.

\clearpage

\section{Determinining which paths are included in a repository}
\label{sec:cross_product}

To expand a templated directory structure into a directory tree, we consider the full cartesian cross product of all placholders' values, and sample a subset of the paths.
The number of paths we sample from the cartesian cross product, $n$, is itself sampled via $n \sim l + Beta(\alpha, \beta)(\min(h, h_{max}) - l)$, where $Beta(\alpha, \beta)$ is the parameterized Beta distribution, $l$ and $h$ are configuration parameters specifying the minimum and maximum global constraints on $n$, and $h_{max}$ is the size of the cartesian cross product, which may be smaller than $h$ sometimes.
In our evaluation, we use $\alpha=1.05$, $\beta = 25$, $l = 15$ and $h=10,000$.
With these parameters the Beta distribution provides a right-skewed distribution that favors file systems with less than $500$ files files, while still allowing for larger repositories.

\section{Using generated Python to define dependent variables}
\label{sec:dependent_variables}

When generating a Python function that might describe a dependent variable, the model is given access to all non-dependent variables, as well as an error term that the model is prompted to integrate into its final computation.

The code block below shows an example function produced by the LLM, representing the dependent variable ``Glucose consumption rate'' in repository \#118. 
Its description, available during generation, is: ``Rate at which glucose is consumed, in grams per liter per hour''. 
The variable is a function of the path variables and independent file variables, described in Appendix~\ref{sec:118_path_variables} and Appendix~\ref{sec:118_file_variables}, respectively.
Line breaks have been introduced for formatting purposes, but the code is otherwise unchanged.

\begin{minted}[fontsize=\small]{python}
import datetime
import math
    
def f_glucose_consumption_rate(variables: dict, error: float) -> float:
    gphase = str(variables["gphase"])
    gtype = str(variables["gtype"])
    date = str(variables["date"])
    tpt = int(variables["tpt"])
    seq_number = int(variables["seq_number"])
    pH = float(variables["pH"])
    temp = str(variables["temp"])
    glucose_conc = str(variables["glucose_conc"])
    oxygen_level = str(variables["oxygen_level"])

    # Base glucose consumption rate: influenced by growth phase, temperature, 
    # glucose concentration, and oxygen
    # In early exponential phase, consumption is high; in late phase, it drops.
    # Temperature affects enzyme kinetics: higher temp increases rate up to an 
    # optimum (~30°C), then declines.
    # Glucose concentration: higher initial concentration may lead to higher 
    # consumption rate, but only up to a point (saturation effect).
    # Oxygen level: aerated conditions support respiration and may slightly 
    # increase glycolytic flux, but in ethanol fermentation, glycolysis dominates.
    # pH: optimal around 4.0–5.0; low pH may inhibit enzymes, reducing 
    # consumption rate.
    # Genotype type: knockouts in glycolytic genes (e.g., pfk1, pyk1) reduce 
    # consumption; overexpression of glycolytic genes may increase it.

    # Convert temperature to numeric (in °C)
    temp_deg = float(temp.replace("°C", "").replace("C", ""))
    
    # Convert glucose concentration to numeric (g/L)
    glucose_conc_g = {"50 g/L": 50, "100 g/L": 100, "200 g/L": 200}[glucose_conc]

    # Growth phase: early = high consumption, mid = peak, late = low
    phase_factor = {"early": 1.5, "mid": 1.0, "late": 0.3}[gphase]

    # Temperature effect: bell-shaped curve around 30°C, Gaussian-like peak at 30°C
    temp_effect = (
        1.0 + 0.1 * (temp_deg - 25) * (1 - math.exp(-0.05 * (temp_deg - 30)**2)))
    temp_effect = max(0.5, min(2.0, temp_effect))  # clamp between 0.5 and 2.0

    # Glucose concentration: higher glucose leads to higher consumption, but 
    # with diminishing returns
    glucose_effect = 0.005 * glucose_conc_g  # linear with glucose concentration, scaled

    # Oxygen level: aerated may slightly increase flux due to better nutrient 
    # delivery and reduced inhibition
    oxygen_map = {"anaerobic": 1.0, "microaerobic": 1.1, "aerated": 1.2}
    oxygen_effect = oxygen_map[oxygen_level]

    # pH effect: optimal around 4.0–5.0; below 4.0 or above 6.0 reduces enzyme activity
    pH_effect = 1.0 + 0.2 * (pH - 4.0) - 0.1 * (pH - 5.0)**2  # parabolic, peaks at 4.5
    pH_effect = max(0.5, min(1.5, pH_effect))  # clamp to reasonable range

    # Genotype type: knockouts reduce consumption; overexpression increases it; 
    # promoter swaps vary
    genotype_map = {"knockout": 0.7, "overexpression": 1.3, "promoter_swap": 1.0}
    genotype_effect = genotype_map[gtype]

    # Time point: consumption rate is highest in early stages and drops over time
    time_effect = 1.0 - (tpt / 48) * 0.5  # linear decay from 0 to 48 hours
    time_effect = max(0.0, min(1.0, time_effect))

    # Base rate (in g/L/h) under optimal conditions
    base_rate = (
        0.8 * phase_factor * temp_effect * glucose_effect 
        * oxygen_effect * pH_effect * genotype_effect * time_effect
    )
    
    # Final rate with added noise (error term)
    # We add error to simulate biological variability
    consumption_rate = base_rate + error

    # Clamp to realistic range (glucose consumption rate cannot be negative or 
    # extremely high)
    # Biological plausibility: typical fermentation consumption rate is 0.1–2.0 g/L/h
    consumption_rate = max(0.0, min(2.0, consumption_rate))

    return consumption_rate
\end{minted}

\section{Populating a file}
\label{sec:populating_a_file}

\begin{algorithm}[ht]
\caption{Populating a file. Given a file seed, a desired file path, an LLM, and distribution parameters to determine the number of rows, we generate the file system, hash the path into a new seed, and sample the independent and dependent variables for that file.}
\label{alg:file}
\begin{algorithmic}
    \Procedure{PopulateFile}{seed, path, LLM, $\mu_\text{rows}$, $\sigma^2_\text{rows}$, $\sigma^2_\text{noise}$}
        \State \textsc{SetGlobalSeed}(seed)
        \State file\_system $\gets$ \textsc{FileSystemSimulator}(LLM)
        \State path\_variables $\gets$ \textsc{ExtractPathVariables}(path, file\_system)
        \State path\_seed $\gets$ \textsc{SHA256}(path)
        \State \textsc{SetGlobalSeed}(path\_seed)
        \State $n_\text{rows} \sim \mathcal{N}(\mu_\text{rows}, \sigma_\text{rows}^2)$
        \State file $\gets\emptyset$
        \State $i \gets 0$
        \For{non\_dependent\_variable $\in$ file\_system}
            \State $\text{file}[i]\sim$ \textsc{SampleNonDependent}(non\_dependent\_variable, $n_\text{rows}$)
            \State $i \gets i + 1$
        \EndFor
        \For{dependent\_variable $\in$ file\_system}
            \State noise $\sim \mathcal{N}(0, \sigma_\text{noise}^2)$
            \State $\text{file}[i]\sim$ \textsc{PopulateDependent}(dependent\_variable, $n_\text{rows}$, path\_variables, noise)
            \State $i \gets i + 1$
        \EndFor
        \State \textbf{return} \textsc{ToFileExtension}(file, path)
    \EndProcedure
\end{algorithmic}
\end{algorithm}

To populate a given file, we follow the procedure in Algorithm~\ref{alg:file}.
Given a random seed that identifies a specific data repository, we set the seed and use the LLM to generate the scientific context, path structures, and variables that define a file system.
We then take the path requested (assuming it is valid for the repository), hash it into an integer using the SHA-256 algorithm, and set the seed again using the hashed value.
We then iterate through the non-dependent and dependent variables, separately, sampling $n_\text{rows}$ data points for each column.
We finally convert the tabular file structure into the path's file extension, and return it.
Since each path in the repository is, by definition, unique, hashing the path into a second random seed guarantees each file is unique, barring hash collisions.
This is also guaranteed due to the path containing variables that may be involved in calculating dependent variables, but this approach also ensures uniqueness amongst variables sampled from named distributions.

\section{Validating Repository Plausibility}
\label{sec:human_study}

As with any simulated benchmark, data realism is critical for any findings to carry weight in actual scientific settings.
To address this, we conduct a new domain-expert validation study covering 6 of the 22 fields used in the benchmark: computer science, biomedical and health, astronomy and space, mechanical and aerospace engineering, biology, psychology and cognitive science. 
We recruit 19 domain experts, all of whom either hold a Ph.D. or are currently pursuing one and conduct research in their current position. 
Each expert is given three generated repositories from their own domain to consider. 
For each repository, they are shown the simulated project title, abstract, and full repository contents through a folder/file interface. 
Table~\ref{tab:human_study} shows the six questions posed to participants.

\begin{table}[h]
    \centering
    \makegapedcells
    \begin{tabular}{l|l}
        \toprule
        \makecell[l]{Question} & Values \\
        \midrule
        \makecell[l]{The folder and file organization looks \\ typical for projects in my field.} & \makecell[l]{Strongly Disagree, Somewhat Disagree, \\ Neither, Somewhat Agree, Strongly Agree} \\
        \makecell[l]{The file naming conventions look \\ realistic for my field.} & \makecell[l]{Strongly Disagree, Somewhat Disagree, \\ Neither, Somewhat Agree, Strongly Agree} \\
        \makecell[l]{The level of messiness/incompleteness \\ looks realistic (e.g. missing values, \\ inconsistent file/directory structure, etc.).} & \makecell[l]{Strongly Disagree, Somewhat Disagree, \\ Neither, Somewhat Agree, Strongly Agree} \\
        \makecell[l]{The column names look plausible for \\ this kind of project.} & \makecell[l]{Strongly Disagree, Somewhat Disagree, \\ Neither, Somewhat Agree, Strongly Agree} \\
        \makecell[l]{The column values are realistic.} & \makecell[l]{Strongly Disagree, Somewhat Disagree, \\ Neither, Somewhat Agree, Strongly Agree} \\
        \makecell[l]{Can you tell this is a simulated repository?} & Yes, Not Sure, No \\
        \bottomrule
    \end{tabular}
    \caption{Questions and response values included in the data plausibility human study.}
    \label{tab:human_study}
\end{table}

Using a one-sided sign test against the null hypothesis that the median rating is neutral-or-worse, we reject the null for all six items ($\alpha=0.05$).
The results provide initial evidence that the generated repositories are plausible enough to provide a reliable signal for evaluation.

\section{Question templates and examples}
\label{sec:question_templates}

Questions cover five categories that get progressively more difficult:

\begin{enumerate}[nosep]
    \item \textbf{Repository Metadata}: these questions do not require reading any tabular data. These questions mainly check if the LLM can recover high-level information like ``Does this repository have a README file?''.
    \item \textbf{File Metadata}: These questions ask high-level questions about an individual file, like ``What is the file extension in this repository''
    \item \textbf{Directory Traversal}: These questions require iterating over some/all subdirectories to answer questions like ``How many files have [PROPERTY] property?''. These questions do not require reading the contents of a file.
    \item \textbf{Univariate Statistics}: These questions ask univariate statistics questions about individual or multiple files, like ``Across files with [PROPERTY] property, what is the average value of the [VARIABLE NAME] variable?
    \item \textbf{Bivariate Statistics}: These are like the univariate statistics category, but they ask the model to calculate the values of bivariate statistics, like the Pearson's correlation between two continuous variables. They can also ask the model whether that statistic is enough to reject a null hypothesis.
    \end{enumerate}

The list below shows an example templated question corresponding to each category/type.

\textbf{Repository Metadata-Readme}
\begin{tcolorbox}[colback=white, colframe=black!30]
    Yes or no, does this repository have a README file?
\end{tcolorbox}

\textbf{Repository Metadata-Title}
\begin{tcolorbox}[colback=white, colframe=black!30]
    Looking at the README file, what is the project title?
\end{tcolorbox}

\textbf{Repository Metadata-Abstract}
\begin{tcolorbox}[colback=white, colframe=black!30]
    Looking at the README file, what is the project abstract?
\end{tcolorbox}

\textbf{File Metadata-Extension}
\begin{tcolorbox}[colback=white, colframe=black!30]
    What is the file extension for the data in this repository?
\end{tcolorbox}

\textbf{File Metadata-Count Rows} (example from repository \#369)
\begin{tcolorbox}[colback=white, colframe=black!30]
    How many rows of data (excluding headers) are in the file: ``intens\_medium-ses\_high-loc\_suburban/rce\_wh/mode\_stem.csv''?
\end{tcolorbox}

\textbf{Directory Traversal-Prefix} (example from repository \#190)
\begin{tcolorbox}[colback=white, colframe=black!30]
    How many files in this repository have the prefix: ``ltypeperiodic/tau2*''?
\end{tcolorbox}

\clearpage

\textbf{Directory Traversal-Condition} (example from repository \#303)
\begin{tcolorbox}[colback=white, colframe=black!30]
    How many files are in this repository where the ``mut\_type'' is ``mim'', the ``date'' is ``2024-05-01'', and the ``seq\_number'' is one of ``1'', ``3'', or ``5''?
\end{tcolorbox}

\textbf{Univariate Statistics-Single File} (example from repository \#458)
\begin{tcolorbox}[colback=white, colframe=black!30]
    In the file ``1-vhigh/meas/int/tool\_AB\_0609/kai.monroe.txt'', only considering rows where the ``joint\_temp'' is less than 52.27, the ``hum\_amb'' is less than 0.3963, and the ``stab'' is ``0'', what is the variance of the ``comm\_latency'' variable?
\end{tcolorbox}

\textbf{Univariate Statistics-Condition} (example from repository \#23)
\begin{tcolorbox}[colback=white, colframe=black!30]
    Only considering files where the ``cond'' is one of ``bal\_bfv'', ``bal\_paillier'', or ``skw\_bfv'' and the ``skw'' is ``bal'', and only considering rows where the ``lr'' is one of ``0.01'', ``0.05'', ``0.1'', or ``0.2'', the ``robustness'' is greater than 99.6, and the ``test\_var'' is at least 8.54, what is the most common value of the ``agg\_weighting'' variable?
\end{tcolorbox}

\textbf{Bivariate Statistics-Statistic} (example from repository \#226)
\begin{tcolorbox}[colback=white, colframe=black!30]
    In the file ``low-neg-vr/yng-5/30m/02\_15\_1\_delay.json'', only considering rows where the ``mean\_latency'' is at least 1500.0, the ``err\_clust\_freq'' is ``scattered'', and the ``ret\_acc'' is one of ``low'' or ``med'', what is the ``Pearson correlation coefficient'' value between the ``err\_rate\_change'' variable and the ``acc\_std'' variable?
\end{tcolorbox}

\textbf{Bivariate Statistics-Hypothesis} (example from repository \#482)
\begin{tcolorbox}[colback=white, colframe=black!30]
    In the file ``1/gaus\_http\_deepflow/1024\_med\_l3/300-08\_25/800\_128\_256.log'', using ``chi-square'' and a p-value of 0.01, can you reject the null hypothesis (yes/no) that there is no relationship between the ``misclass\_rate'' and ``small\_sz\_flag'' variables?
\end{tcolorbox}

\clearpage

\section{Example Paraphrases}
\label{sec:paraphrase_examples}

Below are example paraphrases by the three models used for paraphrasing, along with the original templated version.
For questions with specific file paths, we do not include the path in the templated version given to the paraphrasing LLM.
Instead, we replace the file path with ``\{path\}'', and prompt the LLM to use the variable ``\{path\}'' in its place.
We can then later substitute the path back in to ensure there is no error incurred due to incorrect copying.

\textbf{Univariate Statistics-Condition} (example from repository \#232)
\begin{tcolorbox}[colback=white, colframe=black!30]
    \textbf{Templated} \\
    Only considering files where the "cstr" is "0.0" and the "seq\_number" is "1", and only considering rows where the "t\_to\_ad" is at most 269.3 and the "phase\_coherence" is at least 0.01655, what is the variance of the "phase\_coherence" variable? \\

    \textbf{GPT-OSS 20B} \\
    For the experimental runs with a coupling strength of zero and a trial number of one, and considering only the measurement points where the time to amplitude death is no greater than 269.3 seconds and the phase coherence is at least 0.01655 radians, what is the variance of the phase coherence across those points? \\
    
    \textbf{Gemma 3 27B} \\
    Considering only the data from experiments with no coupling strength and the first experimental trial, what is the variance of the phase coherence among the oscillators for those instances where amplitude death occurred within 269.3 seconds and the oscillators exhibited a phase coherence of at least 0.01655 radians? \\
    
    \textbf{Qwen3 4B Instruct} \\
    Among the first trial runs with no coupling between oscillators, what is the variance in phase coherence when oscillations have not yet died out within 269.3 seconds and the phase coherence is at least 0.01655 radians? \\
\end{tcolorbox}

\textbf{Bivariate Statistics-Hypothesis} (example from repository \#482)
\begin{tcolorbox}[colback=white, colframe=black!30]
    \textbf{Templated} \\
    In the file ``\{path\}'', using ``chi-square'' and a p-value of 0.01, can you reject the null hypothesis (yes/no) that there is no relationship between the ``misclass\_rate'' and ``small\_sz\_flag'' variables? \\
    
    \textbf{Qwen3 4B Instruct} \\
    Based on the experimental data in ``\{path\}'', can we statistically conclude---using a chi-square test at the 0.01 significance level---that there's a meaningful association between the rate at which packets are misclassified due to size anomalies and the presence of any packet below the 64-byte threshold in a flow? \\

    \textbf{Gemma 3 27B} \\
    Considering the data in the file ``\{path\}'', is there a statistically significant relationship between the rate of packets incorrectly classified due to their size and the presence of packets smaller than 64 bytes, using a significance level of 0.01? \\

    \textbf{GPT-OSS 20B} \\
    In the file \{path\}, applying a chi‑square test with a significance level of 0.01, can we reject the null hypothesis that there is no association between the Packet Size Misclassification Rate and whether any packet in the flow is below 64 bytes?\\
\end{tcolorbox}

\section{Questions considered in Evaluation}
\label{sec:questions_in_evaluation}

\begin{table}[h]
    \centering
    \begin{tabular}{l|l|r}
        \toprule
        Question Category & Question Type & Count \\
        \midrule
        Repository Metadata & Readme & 16 \\
        Repository Metadata & Title & 13 \\
        Repository Metadata & Abstract & 16 \\
        File Metadata & Extension & 14 \\
        File Metadata & Count Rows & 82 \\
        Directory Traversal & Prefix & 57 \\
        Directory Traversal & Condition & 53 \\
        Univariate Statistics & Single File  & 55 \\
        Univariate Statistics & Condition & 61 \\
        Bivariate Statistics & Statistic & 64 \\
        Bivariate Statistics & Hypothesis & 69 \\
        \bottomrule
    \end{tabular}
    \caption{The number of sampled questions used per question category and question type in the 500 questions included in the evaluation.}
    \label{tab:questions}
\end{table}

We evaluate all models on a sample of 500 questions, split into 361 answerable questions (72.2\%) and 139 unanswerable questions (27.8\%).
For each question, we sample model responses on the templated variant and the three paraphrase variants.
This sample size is guided by a two-sided paired t-test ($\alpha=0.01$, $1-\beta=0.95$, effect size $d=0.2$), where the null hypothesis is that accuracy of all models comes from the same distribution.

Table~\ref{tab:questions} shows counts by question category and type corresponding to the 500 questions used in the evaluation.
For questions that have stochasticity, because they refer to a sampled path or they contain sampled filtering conditions, we generate five seeded examples per repository. 
This is why the last seven question types have about 5 times as many samples as the first four question types. 

In our evaluation sample, the counts of the file extensions corresponding to each question's repository are: 76 .csv, 94 .json, 83 .jsonl, 77 .log, 83 .txt, and 87 .xlsx.

\section{Evaluation details}
\label{sec:eval_details}

Since GPT-5.4 and Claude Opus 4.6 both provide Model Context Protocol (MCP) compatible interfaces, we provide their APIs access to the tools and delegate managing the agentic state- and action-spaces to them.
We set the reasoning effort hyperparameter to ``medium'' for both models to strike a balance between accuracy and token/tool efficiency.
For the open-weight models, we run a ReAct-like framework \citep{yao2023react} using the smolagents library \citep{smolagents}.
We use recommended decoding hyperparemeters where specified in the models' documentation, and defer to setting the temperature $t=1.0$ otherwise.

The data is made available to the model as an MCP server that presents both both directory and file reading functions.
Each function accepts an \verb|id| parameter, corresponding to the random seed that identifies and defines a data repository.
We also provide the LLM with a Python interpreter MCP server to help programatically interact with the data.
It has common data science libraries, like ``numpy'', ``pandas'', and ``scikit-learn''.
The data-reading tools are also explicitly made available inside the Python interpreter, so an LLM can interact with directories and files without ever loading them into the context window.
Overall, the two MCP servers support the following functions:
\begin{enumerate}[nosep]
    \item \verb|list_directory(id,prefix,depth)| - lists the subdirectories of the directory beginning with the prefix. It accepts wildcards (``*'' and ``?''), similar to the bash \verb|ls| command.
    \item \verb|read_text_file(id,path,head,tail)| - returns the contents of a file at the specified path, with the options head and tail truncating the amount of content returned like the bash \verb|head| and \verb|tail| commands.
    \item \verb|read_binary_file(id,path)| - returns the Base64 encoded contents of the file, along with the file's MIME type.
    \item \verb|run_python_code(code)| - accepts a block of Python code and executes it, subject to a 60s time limit and a 512MB compute limit, returning either the output or the error produced during execution.
\end{enumerate}

We prompt all LLMs to return a JSON structure to return their answer, simply mapping the key answer to their answer, like: \texttt{\{"answer": "..."\}}.
This is helpful in cases where models natural perform Chain-of-Thought reasoning, as it separates the model's reasoning from its final answer.
If the model does not provide valid JSON in its output ($<3\%$ of responses), we consider the entire final response as the answer.
To measure unanswerability, every question prompts the model to provide ``not possible'' as its answer if it deems the question unanswerable.

Grading model responses is entirely deterministic
When grading LLM responses to questions, we use the following procedure:

\begin{itemize}[nosep]
    \item If the question's answer is \textbf{categorical} and finite (not open-ended), we mark the LLM response as correct if it contains the correct answer and does not contain any of the other choices, normalizing everything to lower case.
    \item If the question's answer is \textbf{discrete integer}, we attempt to cast the LLM's response to integer and measure whether there is an exact match. If that is not possible, perhaps because the LLM has included units of measurement in the answer (e.g. ``10cm''), we use a regular expression to extract contiguous numeric characters, cast those to integer, and then evaluate an exact match.
    \item If the question's answer is \textbf{continuous}, we consider the number of significant digits used in the question. Every question contains a preamble that stipulates that the LLM should ``use $x$ number of significant figures, if the answer is continuous.'' This values is randomly chosen from $\{2,3,4\}$. If the answer can be cast to float, we use that. Otherwise, we attempt to extract contiguous numeric characters (including ``.'' since it is continuous) and try to them that to float. Once we have an answer extracted, we consider \emph{one less} significant digit than was requested in the question, and see if there is an exact match. For example, if the question asks for $3$ significant digits, the correct answer is $1.234$, and the model's response is $1.235$, the response would be marked correct since three significant digits only considers whether the response contains $1.23$.
\end{itemize}

\section{Example LLM responses}
\label{sec:example_response}

Both of the following examples are responses to the same question.
Appendix~\ref{sec:gem} shows an example trajectory from Gemma 3 27B it, and  Appendix~\ref{sec:gpt} shows an example trajectory from GPT-5.4.
In the GPT-5.4 example, the model response is shown as \verb|<encrypted>|, since the contents of the reasoning chain are not provided by by the OpenAI interface.
The full output has been truncated for formatting purposes, and to skip verbose output that is not related to answering the question.

The correct answer to the question is 163, answered correctly by GPT-5.4, and answered incorrectly by Gemma 3 27B it.
By examining the correct and incorrect responses, this example demonstrates that LLM strategies that prioritize solving problems using code are more likely to answer correctly than LLMs that rely on their context window. 
Gemma 3 27B it correctly navigates to the directory and opens the file, but it then uses the context window to load the file, and provides the wrong answer.
In contrast, GPT-5.4 loads only the first 40 lines of the file, recognizes that it can load the file using Python, and then uses the Python interpreter tool to load the file and print the answer.

\subsection{Gemma 3 27B it}
\label{sec:gem}

\begin{tcolorbox}[colback=white, colframe=black!30]

\textbf{Question}
\begin{minted}[fontsize=\small]{text}
<TOOL INSTRUCTIONS>

This question is about filesystem #23. Any calls to the scientific data repository
should specify `id=23`.

How many data rows, excluding the header, are in the file located at
'cond=skw_paillier_ns=0.05/ncl=50-arch=mlp2/encr=bfv/rounds=10-skw=skw-2.json'?

When you have your answer ready, return a JSON object with a single key, "answer",
mapping to your answer. If the question doesn't have an answer, reply "not 
possible" as your answer, like this: `{"answer": "not possible"}`.
\end{minted}
\end{tcolorbox}

\begin{tcolorbox}[colback=white, colframe=black!30]

\textbf{[Step 1] Model Response}
\begin{minted}[fontsize=\small]{text}
We need to count data rows, excluding the header, in a specified JSON file within
some repository. Directory structure indicates repository id=23. Need to list the
directory, then read the file to understand its contents. We'll call list_directory
with id=23 prefix pointing to /cond=skw_paillier_ns=0.05/ncl=50-arch=mlp2/encr=bfv/
rounds=10-skw=skw-2.json? Wait the file location is a path relative? The repository
likely root at /? We need to confirm file path. Use list_directory to see root. 
Then navigate to path. Let’s list root depth maybe multiple.
\end{minted}

\textbf{Tool Call}
\begin{minted}[fontsize=\small]{json}
{
  "tool_name": "read_text_file",
  "arguments": {"id": 23, "prefix": "/*", "depth": 1}
}
\end{minted}

\textbf{Observations}
\begin{minted}[fontsize=\small]{json}
{
  "status": "success", 
  "paths": [
    "cond=bal_bfv_ns=0.01", 
    "cond=bal_bfv_ns=0.05"
    //...
  ] 
}
\end{minted}
\end{tcolorbox}

\begin{tcolorbox}[colback=white, colframe=black!30]
\textbf{[Step 2] Model Response}

\begin{minted}[fontsize=\small]{text}
We need to navigate to folder cond=skw_paillier_ns=0.05 then subfolder 
ncl=50-arch=mlp2/encr=bfv/rounds=10-skw=skw-2.json. Use list_directory depth 2 to
list contents of that folder to confirm existence.

First path: /cond=skw_paillier_ns=0.05/ncl=50-arch=mlp2/encr=bfv/rounds=10-
skw=skw-2.json? Wait file extension .json; not a folder. Actually file path 
indicates nested directories: cond=skw_paillier_ns=0.05 / ncl=50-arch=mlp2 / 
encr=bfv / file 'rounds=10-skw=skw-2.json'.

But path shows folder encr=bfv and file inside maybe. Let's list at 
/cond=skw_paillier_ns=0.05/ncl=50-arch=mlp2.
\end{minted}

\textbf{Tool Call}
\begin{minted}[fontsize=\small]{json}
{
  "tool_name": "list_directory",
  "arguments": {
    "id":23,
    "prefix":"/cond=skw_paillier_ns=0.05/ncl=50-arch=mlp2",
    "depth":1
  }
}
\end{minted}

\textbf{Observations}
\begin{minted}[fontsize=\small]{json}
{"status": "success", "paths": ["encr=bfv"]}
\end{minted}
\end{tcolorbox}

In steps 3-6, the LLM navigates to the correct folder and confirms the file exists.
We have omitted these steps for brevity, but they exhibit similar patterns to steps 1 and 2 above.

\begin{tcolorbox}[colback=white, colframe=black!30]
\textbf{[Step 7] Model Response}

\begin{minted}[fontsize=\small]{text}
We need to open file /cond=skw_paillier_ns=0.05/ncl=50-arch=mlp2/encr=bfv/
rounds=10-skw=skw-2.json.
\end{minted}

\textbf{Tool Call}
\begin{minted}[fontsize=\small]{json}
{
  "tool_name": "list_directory",
  "arguments": {
    "id":23,
    "path":"cond=skw_paillier_ns=0.05/.../rounds=10-skw=skw-2.json",
    "head": null,
    "tail": null
  }
}
\end{minted}

\textbf{Observations}
\begin{minted}[fontsize=\small]{json}
{"status": "success", "file_content": "[\n  {\n    \"lr\":\"0.05\",\n..."}
\end{minted}
\end{tcolorbox}

\begin{tcolorbox}[colback=white, colframe=black!30]
\textbf{[Step 8] Model Response}

\begin{minted}[fontsize=\small]{text}
We need to count the number of rows in the file. Looking at the content of the file,
the file has 157 rows.

The final answer is 157 rows.
\end{minted}

\textbf{Tool Call}
\begin{minted}[fontsize=\small]{json}
{
  "tool_name": "final_answer",
  "arguments": {"answer":157}
}
\end{minted}
\end{tcolorbox}

\subsection{GPT-5.4}
\label{sec:gpt}

\begin{tcolorbox}[colback=white, colframe=black!30]

\textbf{Question}
\begin{minted}[fontsize=\small]{text}
<TOOL INSTRUCTIONS>

This question is about filesystem #23. Any calls to the scientific data repository
should specify `id=23`.

How many data rows, excluding the header, are in the file located at
'cond=skw_paillier_ns=0.05/ncl=50-arch=mlp2/encr=bfv/rounds=10-skw=skw-2.json'?

When you have your answer ready, return a JSON object with a single key, "answer",
mapping to your answer. If the question doesn't have an answer, reply "not 
possible" as your answer, like this: `{"answer": "not possible"}`.
\end{minted}
\end{tcolorbox}

\begin{tcolorbox}[colback=white, colframe=black!30]

\textbf{[Step 1] Model Response}
\begin{minted}[fontsize=\small]{text}
<ENCRYPTED>
\end{minted}

\textbf{Tool Call}
\begin{minted}[fontsize=\small]{json}
{
  "tool_name": "read_text_file",
  "arguments": {
    "id":23,
    "path":"cond=skw_paillier_ns=0.05/.../rounds=10-skw=skw-2.json",
    "head":40,
    "tail":null
  }
}
\end{minted}

\textbf{Observations}
\begin{minted}[fontsize=\small]{json}
{"status": "success", "file_content": "[\n  {\n    \"lr\":\"0.05\",\n..."}
\end{minted}
\end{tcolorbox}

\begin{tcolorbox}[colback=white, colframe=black!30]
\textbf{[Step 2] Model Response}

\begin{minted}[fontsize=\small]{text}
<ENCRYPTED>
\end{minted}

\textbf{Tool Call}
\begin{minted}[fontsize=\small]{json}
{
  "tool_name": "run_python_code",
  "arguments": {
    "code": "import json\ntext = read_text_file(id=23, path='cond=skw_pai..."
}
\end{minted}

\textbf{Observations}
\begin{minted}[fontsize=\small]{json}
{"status": "success", "output": "163\n", "error": null}
\end{minted}
\end{tcolorbox}

\begin{tcolorbox}[colback=white, colframe=black!30]
\textbf{[Step 3] Model Response}

\begin{minted}[fontsize=\small]{text}
{"answer": 163}
\end{minted}
\end{tcolorbox}

\clearpage

\section{Paraphrase Validity Audit}
\label{sec:paraphrase_validity}

\begin{table}[h]
    \centering
    \begin{tabular}{l|rrr}
    \toprule
         & Semantically Equivalent & Unambiguous & Fluent \\
        \midrule
        GPT-OSS 20B & 96\%  & 98\%  & 100\% \\
        Gemma 3 27B it & 62\%  & 68\%  & 96\% \\
        Qwen3 4B Instruct & 78\%  & 80\%  & 94\% \\
        \bottomrule
    \end{tabular}
    \caption{Results of the paraphrase validity audit per paraphrase model. Each value shows the proportion of the 50 sampled question paraphrases that matches the column title, when compared against the templated original.}
    \label{tab:paraphrase_validity_audit}
\end{table}

The paraphrase module is intended to add warranted difficulty by returning the unnatural, templated question back to a naturalistic form that might be asked by a scientist.
Looking at model accuracy alone does not guarantee that the added difficulty is not introduced by some other factor.
To assess the quality of the paraphrases, we manually inspect a random subset of 50 template/paraphrase question pairs, across all three paraphrase models.
Namely, we evaluate whether every paraphrase is semantically equivalent, unambiguous with respect to the template, and fluent.

Table~\ref{tab:paraphrase_validity_audit} shows the results of this audit. 
Although all models produce fluent paraphrases, only GPT-OSS 20B produces paraphrases that are both semantically equivalent and unambiguous.
These results suggest that GPT-OSS 20B paraphrases add the desired difficulty mimicking realistic question forms, while paraphrases produced by Gemma 3 27B it and Qwen3 4B Instruct can be seen more as a robustness check whose downstream results should be taken with a grain of salt.

\end{document}